\documentclass[journal]{IEEEtran}

\usepackage{color}
\usepackage{times,amsmath,epsfig}
\usepackage{amssymb}
\usepackage{amsthm}
\usepackage{subcaption}
\usepackage{cite}
\usepackage{hyperref}
\usepackage{url}
\usepackage[capitalise]{cleveref}
\usepackage[ruled,linesnumbered]{algorithm2e}
\usepackage[dvipsnames,svgnames]{xcolor}
\usepackage{tikz, pgfplots}
\usepackage{moresize}
\usepackage{array}
\usepackage{comment}
\usepackage{booktabs}
\usepackage{tabularx}
\usepackage{multirow}
\usepackage{enumitem}

\usetikzlibrary{shapes,arrows}
\pgfplotsset{compat=1.17}
\pgfplotstableset{col sep=comma}
\usepgfplotslibrary{groupplots}

\newtheorem{theorem}{Theorem}
\newtheorem{definition}{Definition}
\newtheorem{proposition}{Proposition}

\input{my_styles.sty}

\newcommand \victor[1]{{\color{teal}[Victor: #1]}}

\newcommand{\alna}[1]{\begin{alignat}{3}&#1&\end{alignat}}

\def \hArw {{\hat{\bbA}_{\rm \scriptscriptstyle rw}}}
\def \hArws {{\hat{\bbA}^*_{\rm \scriptscriptstyle rw}}}

\pgfplotsset{compat=1.17}
\pgfplotstableset{col sep=semicolon}
\usepgfplotslibrary{fillbetween}
\tikzset{every mark/.append style={scale=1.5, solid}, font=\small}
\pgfplotsset{
    width=1.05\textwidth,
    height=5.5cm,
    legend style={
        font=\ssmall ,  
        inner xsep=1pt,
        inner ysep=1pt,
        nodes={inner sep=1pt}},
    legend cell align=left,
    every axis/.append style={line width=.5pt},
 	every axis plot/.append style={line width=1.5pt},
 	every axis y label/.append style={yshift=-4pt}
}
\usepgfplotslibrary{external}
\begin{document}

\title{Adapting to Heterophilic Graph Data with Structure-Guided Neighbor Discovery}
%
%
%

\author{Victor M. Tenorio,~\IEEEmembership{Student Member,~IEEE,}
        Madeline Navarro,~\IEEEmembership{Student Member,~IEEE,}
        Samuel Rey,~\IEEEmembership{Member,~IEEE,}
        Santiago Segarra,~\IEEEmembership{Senior Member,~IEEE,}
        and~Antonio G. Marques,~\IEEEmembership{Senior Member,~IEEE}
\thanks{V. M. Tenorio, S. Rey, and A. G. Marques are with the Department of Signal Theory and Communications, King Juan Carlos University, Madrid, Spain. Emails: \{victor.tenorio,samuel.rey.escudero,antonio.garcia.marques\}@urjc.es}
\thanks{M. Navarro and S. Segarra are with the Department of Electrical and Computer Engineering, Rice University, Houston, TX, US. Emails: \{nav,segarra\}@rice.edu}
\thanks{This work was partially supported by NSF under award CCF-2340481; by the Spanish AEI (10.13039/501100011033) under Grants PID2022-136887NBI00 and FPU20/05554, and the Community of Madrid within the ELLIS Unit Madrid framework and the IDEA-CM (TEC-2024/COM-89) and CAM-URJC F861 and F1180 (CP2301) grants.}
\thanks{Research was sponsored by the Army Research Office and was accomplished under Grant Number W911NF-17-S-0002. The views and conclusions contained in this document are those of the authors and should not be interpreted as representing the official policies, either expressed or implied, of the Army Research Office or the U.S. Army or the U.S. Government. The U.S. Government is authorized to reproduce and distribute reprints for Government purposes notwithstanding any copyright notation herein.}}

%

\maketitle

\begin{abstract}
Graph Neural Networks (GNNs) often struggle with heterophilic data, where connected nodes may have dissimilar labels, as they typically assume homophily and rely on local message passing. To address this, we propose creating alternative graph structures by linking nodes with similar structural attributes (e.g., role-based or global), thereby fostering higher label homophily on these new graphs. We theoretically prove that GNN performance can be improved by utilizing graphs with fewer false positive edges (connections between nodes of different classes) and that considering multiple graph views increases the likelihood of finding such beneficial structures. Building on these insights, we introduce Structure-Guided GNN (SG-GNN), an architecture that processes the original graph alongside the newly created structural graphs, adaptively learning to weigh their contributions. Extensive experiments on various benchmark datasets, particularly those with heterophilic characteristics, demonstrate that our SG-GNN achieves state-of-the-art or highly competitive performance, highlighting the efficacy of exploiting structural information to guide GNNs.
\end{abstract}


%
\IEEEpeerreviewmaketitle

\section{Introduction}
\label{S:intro}
\IEEEPARstart{G}{raph} neural networks (GNNs) have demonstrated remarkable performance in processing graph-structured data by leveraging local neighborhood information~\cite{wu2021comprehensive, bronstein2017geometric}.
For node classification or regression, the graph structure is traditionally assumed to indicate which nodes should share similar representations or be treated similarly by a GNN.
This holds true in many real-world scenarios; for example, temperature readings on a graph discretizing a geographical region often exhibit smoothness, where nearby sensors record similar temperatures~\cite{bronstein2017geometric}. Similarly, convolutional neural networks~\cite{li2022cnn} can be viewed as a special case of GNNs operating on grid-like graphs, where the inherent smoothness of natural images (i.e., adjacent pixels often having similar values) aligns well with the local filtering operations of GNNs~\cite{bronstein2017geometric,tenorio24iognn}.
However, it is increasingly recognized that while the observed graph topology may contain valuable information, its edges may not directly indicate node similarity~\cite{zhu2020beyond,ma2022homophily}.
In such settings, directly employing the observed graph as input to a GNN can be suboptimal and may even hinder performance.

A prominent example of this is heterophily, where connected nodes may be dissimilar or similar nodes may be distant~\cite{zhu2020beyond, pei2020geom}.
Traditional GNNs typically rely on local, low-pass filters, hence struggling with such data~\cite{nt2020revisiting}.
To address this, several alternative methods have been proposed to incorporate non-local information.
Some propose new architectures that are well suited for heterophilic data~\cite{abu2019mixhop,bo2021beyond,lim2021large,chien2021adaptive}, although these models may decrease interpretability due to their added complexity, and their performance may depend on the data being heterophilic, which may not be known beforehand~\cite{chen2020simple,gasteiger2019diffusion,gasteiger2019predict}.
Alternatively, GNNs that are not constrained to learn low-pass representations can adapt to both homophilic and heterophilic data.
However, these models may resort to high powers of the adjacency matrix (leading to potential numerical instability) and require more parameters to allow for the necessary flexibility~\cite{ruiz2021graph,gama2020stability}, so while they are demonstrably effective at learning from heterophilic data, they may require a more careful optimization and be outperformed by simpler architectures on homophilic datasets~\cite{rey2024redesigning}. 

This problem motivates neighbor discovery for node-level predictions~\cite{suresh2021breaking,wu2024beyond}.
Instead of altering GNN architectures, we may instead consider creating graphs more suited to the problem at hand.
An alternative graph may have a more explicit relationship with node labels, allowing us to either leverage the inherent low-pass filtering of GNNs if node labels are homophilic on the new structure~\cite{tenorio2024structure} or reduce complexity for GNNs designed to adapt to heterophilic data~\cite{ruiz2021graph}.
Deriving a new graph may involve using entirely new information or incorporating the original structure without directly using its potentially misleading edges.
While heuristically altering the original graph can be effective, systematically exploiting known structural relationships offers a robust and convenient way to mitigate data dependencies and promote desired behaviors, independent of label noise.

To this end, we propose a structure-based neighbor discovery approach to design alternative graph structures on which node labels exhibit greater homophily.
Our methodology is inspired by the empirical observation that, in real-world graph data, nodes with similar structural attributes often share analogous labels.
Thus, for a given node-level task, we construct new graphs based on sets of structural attributes computed for each node from the original graph.
While our framework is flexible to various structural attributes, we exemplify its application using two primary categories: \emph{role-based} attributes, which encode local connectivity patterns of nodes~\cite{guo20role, tenorio2024recovering}, and \emph{global} attributes, which relate a node to the entire graph structure~\cite{newman2018networks}.
Recognizing that the optimal graph choice is often unknown a priori and that both the original and newly derived graphs can be informative, we further introduce the \emph{Structure-Guided GNN (SG-GNN)}. This architecture processes multiple graphs, including the original and the new structurally informed ones, and adaptively learns to emphasize the most relevant graphs for the task. This adaptive mechanism also enhances interpretability by revealing which graph structures contribute most to performance.

In particular, we make the following contributions:
\begin{itemize}[align=parleft, leftmargin=*]
    \item
        We introduce a structure-based neighbor discovery approach to construct new graphs from node-wise structural attributes of the original graph. This method is motivated by our empirical analysis of real-world graph data, which reveals that nodes possessing similar structural features often share the same labels.

    \item
        We theoretically prove that GNN-based node classification can be improved by utilizing a graph with fewer edges connecting nodes of different classes. We further show that incorporating multiple graph views increases the likelihood of identifying such an improved graph.

    \item
        Motivated by our theoretical findings, we propose SG-GNN. This model integrates the original graph with a set of structure-based graphs within a GNN layer, employing adaptively learned weights to aggregate their contributions and determine the most relevant graph structures.

\end{itemize}

\subsection{Related Works}
\label{Ss:related}

The challenge of applying GNNs to heterophilic data has spurred significant research, primarily along two main avenues: designing GNN architectures suitable for non-homophilic data and modifying the input graph structure. 

\noindent
\textbf{GNN Architectures for Heterophily.}
Many efforts have focused on developing novel GNN architectures tailored for heterophily. 
Some propose architectures that specifically incorporate higher-order neighborhoods or signed messages, which may not generalize well when data is not explicitly either homophilic or heterophilic~\cite{gasteiger2019predict,gasteiger2019diffusion,chen2020simple}.
Other approaches aim for broader applicability by allowing nodes to adaptively incorporate information from neighbors near, far, and in between~\cite{zhu2020beyond,abu2019mixhop,bo2021beyond,chien2021adaptive,lim2021large,rey2024redesigning,ruiz2021graph,gama2020stability}.
While powerful, these general-purpose models can become overly complex, increasing the risk of overfitting and often suffering from a lack of interpretability regarding how they adapt to different homophily levels.
A subset of these methods implicitly or explicitly exploit structural information. 
For instance, some models incorporate centrality scores like PageRank into their aggregation schemes or attention mechanisms~\cite{gasteiger2019predict}. However, these often embed such structural cues within complex, end-to-end learned models, where the direct impact or interpretation of specific structural attributes can be obscured.

\noindent
\textbf{Neighbor Discovery.}
An alternative line of work, often termed graph rewriting or neighbor discovery, focuses on modifying or creating new graph topologies on which node labels exhibit greater homophily, allowing standard GNNs to perform more effectively~\cite{wu2024homophily,jin2020graph}.
Common strategies include heuristically modifying the original graph by adding or removing edges based on feature similarity or label information (if available during a pre-processing stage)~\cite{chen2020measuring}. 
Some approaches discard the original graph entirely and construct a new one, for example, by building a $k$-Nearest Neighbors ($k$-NN) graph based on node features~\cite{jin2021similarity,li2024seeking}.
Other methods use the original graph structure to inform the discovery of new neighborhoods but do not directly use the original edges in the GNN message passing.
Examples include learning node embeddings that capture structural roles or similarities, and then constructing a new graph from these embeddings~\cite{wang2022unbiased,suresh2021breaking}.
While these works align with our intuition about exploiting graph structure indirectly, the use of uninterpretable and computationally expensive embeddings such as Struc2Vec~\cite{ribeiro2017struc2vec} can hinder their applicability and may not allow us to identify which characteristics of the original graph structure relate to the learning task.

The short conference publication~\cite{tenorio2024structure} previously introduced a similar structure-based neighbor discovery approach. This paper significantly extends the initial concepts in~\cite{tenorio2024structure} by incorporating a theoretical error characterization for the GNN, developing a multi-layer architecture for the proposed model, providing a theoretical analysis of the benefits derived from aggregating information across multiple graphs, and presenting a substantially expanded set of new experiments.

\section{Preliminaries}
\label{S:prelim}

This section introduces the basics of GNNs for node classification, followed by essential background on homophilic graph data and useful measurements of graph structural properties.

\subsection{Graph Neural Networks for Node Classification}
\label{Ss:gnn}

Consider an undirected graph $\ccalG = (\ccalV, \ccalE)$ with a set of $N = |\ccalV|$ nodes $\ccalV$ and an edge set $\ccalE \subseteq \ccalV\times \ccalV$.
Each node $i \in \ccalV$ is associated with a neighborhood $\ccalN(i) := \{ j\in\ccalV \,:\, (i,j)\in\ccalE \}$.
The connectivity of the graph  is represented by its adjacency matrix $\bbA \in \reals^{N \times N}$, where $A_{ij}\neq 0$ if and only if $(j,i)\in\ccalE$.
Graph learning tasks frequently rely on normalized versions of $\bbA$.
Relevant examples include the symmetrically normalized adjacency matrix with self-loops, $\hbA = \hbD^{-1/2}(\bbA + \bbI)\hbD^{-1/2}$ where $\hbD = \diag((\bbA+\bbI)\bbone)$ typically used in GCNs~\cite{kipf17gnns}, and the row-normalized (or random walk) adjacency matrix, $\hArw =  \hbD^{-1} (\bbA + \bbI)$.
Our primary focus is on the task of node classification~\cite{kipf17gnns}. 
In this setting, each node is equipped with an $M$-dimensional feature vector, collectively represented by the feature matrix $\bbX \in \reals^{N\times M}$, and assigned a class label from $\bby \in \{1,\dots,C\}^N$, where $C$ is the total number of classes.
Given the node features $\bbX$, the graph structure $\bbA$, and the training labels $\bby_{\rm train}$, which are a subset of the full label vector $\bby = [\bby_{\rm train}^\top, \bby_{\rm test}^\top]^\top$, the goal is to estimate the unknown labels $\bby_{\rm test}$.
To this end, we learn a non-linear parametric mapping, $\Phi(\cdot; \bbA, \bbTheta):\reals^{N\times M} \rightarrow \{1,\dots,C\}^N$, typically implemented as a GNN, whose output $\hat{\bby} = \Phi(\bbX; \bbA, \bbTheta)$ predicts the true node labels.
The parameters $\bbTheta$ are learned by minimizing a loss function over the training set according to
\alna{
    \min_{\bbTheta} ~ \ccalL(\bby_{\rm train}, \Phi(\bbX; \bbA, \bbTheta)),
\label{eq:opt_prob}}
where $\ccalL$ is a loss function measuring classification error, such as cross-entropy loss.
The GNN $\Phi$ is typically implemented as a stack of $L$ layers. The transformation at the $\ell$-th layer, producing hidden representations $\bbH^{(\ell+1)}$ from $\bbH^{(\ell)}$ (with $\bbH^{(0)} = \bbX$), can be generally expressed as
\alna{
    \bbH^{(\ell+1)}
    &~=~&
    \phi\left(
        \bbH^{(\ell)}; \bbA, \bbTheta^{(\ell)}
    \right),
\label{eq:gcn_layer}}
where $\bbTheta^{(\ell)}$ denotes the learnable parameters of layer $\ell$, and $\phi$ represents the function implemented by a single GNN layer.
We denote the collection of parameters of the GNN as $\bbTheta = \{ \bbTheta^{(\ell)} \}_{\ell = 1}^L$).
Examples of such layer functions include those from Graph Convolutional Networks (GCNs)~\cite{kipf17gnns} or Graph Attention Networks (GATs)~\cite{velickovic2018graph}.
Most GNN layers perform a variation of a graph filtering operation, where node features are aggregated from their local neighborhoods, followed by a non-linear activation function~\cite{rey2024redesigning}.
For example, a GNN with a bank of graph filters (FBGNN)~\cite{ruiz2021graph,gama2020stability} reads as
\alna{
    \bbH^{(\ell+1)}
    &~=~&
    \sigma \left( \sum_{s=0}^{S-1}
        \hbA^{s} \bbH^{(\ell)} \bbTheta^{(\ell)}_s
    \right),
\label{eq:gcn_gf_layer}}
where $S$ is the order of the graph filter, $\sigma$ is a pointwise nonlinearity and every power of the adjacency matrix is weighted by a different $\bbTheta^{(\ell)}_s$. 

\subsection{Homophilic Graph Data}
\label{Ss:homophily}

The foundation of many graph learning techniques, particularly classical GNNs, rests on the principle of homophily, i.e., the tendency for nodes with similar attributes or labels to be connected in a graph~\cite{mcpherson2001birds}.
This principle directly motivates the use of GNNs for node-level predictions. 
Many GNNs function as low-pass graph filters, meaning their operations encourage the output node representations to be smooth across the graph structure; nodes connected by an edge are pushed towards having similar representations~\cite{kipf17gnns, wu2019simplifying}.
Consequently, having graph data that exhibits homophily with respect to the task labels is valuable for maximizing GNN performance~\cite{zhu2020beyond, ma2022homophily}.
The implications of this for GNN outputs, particularly regarding smoothness, are further discussed in Section~\ref{S:smoothness}.

Given its importance, we consider several quantitative measures of homophily.
First, \emph{edge homophily} (often termed homophily ratio) directly measures the fraction of edges in the graph that connect nodes of the same class~\cite{pei2020geom, zhu2020beyond} as
\alna{
    h_{\rm edge}
    &~=~&
    \frac{ | \{(i,j)\in\ccalE \,:\, y_i=y_j\} | }{ |\ccalE| }.
\label{eq:edge_hom}}
A higher $h_{\rm edge}$ (closer to 1) indicates a graph structure where connections predominantly occur between nodes sharing the same label.

Second, \emph{node homophily} provides a localized, per-node measure of this tendency. For a given node $i$, it is the fraction of its neighbors $\ccalN(i)$ that belong to the same class as node $i$~\cite{pei2020geom, luan2021heterophily}, and is computed as
\alna{
    h_{\rm node}(i)
    &~=~&
    \frac{ | \{j \in \ccalN(i) \, : \, y_j=y_i\} | }{ |\ccalN(i)| }.
\label{eq:node_hom}}
Thus, $h_{\rm node}(i)$ reflects how homophilic the immediate neighborhood of node $i$ is with respect to its own class label $y_i$.

A closely related concept from GSP is \emph{total variation (TV)}, which quantifies the smoothness of a signal (such as node features or labels) over a graph~\cite{sandryhaila2014discrete}. For a given matrix of node features $\bbX\in\reals^{N\times M}$, its TV on a graph with normalized adjacency matrix $\hbA$ is defined as
\alna{
    TV(\bbX)
    &~=~&
    \frac{1}{M}
    \| \bbX - \hbA\bbX \|_1,
\label{eq:tv}}
This measures the average difference between a node's features and the features of its neighbors. We are particularly interested in the TV of the node labels $\bby$. Let $\bbY \in \{0,1\}^{N\times C}$ be the one-hot representation of class labels, where $Y_{ic}=1$ if node $i$ belongs to class $c \in \{1,\dots,C\}$. A lower $TV(\bbY)$ implies that connected nodes tend to have similar labels, aligning with the notion of high homophily.

\subsection{Structural Attributes on Graphs}
\label{Ss:structure}

While the input graph $\ccalG$ offers connectivity data, its direct edges may not always best capture node similarity for classification. We thus leverage well-founded topological metrics from graph theory and network science to characterize each node's structural properties, providing alternative perspectives on node relationships beyond literal connections.

\emph{Role-based structural attributes} aim to distinguish nodes by their local connectivity patterns, such as degree or triangle counts~\cite{henderson2012rolx,guo20role}. For instance, in a university network, students and faculty often exhibit distinct local patterns reflecting their roles~\cite{pei2020geom}. Such structural role understanding has a foundation in node embeddings~\cite{ribeiro2017struc2vec} and has informed GNN design, often via role-aware embeddings~\cite{suresh2021breaking}.

\emph{Global structural attributes}, conversely, contextualize nodes by their position within the entire graph. Centrality measures are prime examples, quantifying node influence or importance network-wide~\cite{newman2018networks}. These metrics have a long history in identifying key nodes in social or infrastructure networks~\cite{freeman1977set} and have been integrated into GNNs to capture broader structural context~\cite{gasteiger2019predict}.

We will later show that real-world graph data often exhibits significant homophily with respect to these structural attributes. 
While the framework of using structural attributes is broad, we focus on these specific role-based and global categories as illustrative examples due to their interpretability and computational efficiency, and, as we will demonstrate, their effectiveness in enhancing homophily on real-world data. For a detailed list of the specific attributes we employ, refer to Appendix~\ref{app:struc_features}.

\begin{table*}[t!]
    \centering
    \begin{tabular}{lcccccccc}
    \toprule
    \multirow{2}{*}{\textbf{Dataset}}& \multicolumn{4}{c}{$TV (\bbY)$} & \multicolumn{4}{c}{Edge homophily $h_{\rm edge}$} \\
    \cmidrule(lr){2-5}\cmidrule(lr){6-9}
     &  $\ccalG$ &  $\ccalG^{\rm nn}_{k,{\rm feat}}$ &  $\ccalG^{\rm nn}_{k,{\rm role}}$ & $\ccalG^{\rm nn}_{k,{\rm glob}}$ &  $\ccalG$ &  $\ccalG^{\rm nn}_{k,{\rm feat}}$ &  $\ccalG^{\rm nn}_{k,{\rm role}}$ & $\ccalG^{\rm nn}_{k,{\rm glob}}$ \\
    \midrule
    Texas & 0.2488 & 0.2065 & 0.1938 & 0.1977 & 0.0609 & 0.5597 & 0.6141 & 0.5796 \\
    Wisconsin & 0.2399 & 0.1982 & 0.2143 & 0.2148 & 0.1778 & 0.5656 & 0.4552 & 0.4527 \\
    Cornell & 0.2453 & 0.2012 & 0.2097 & 0.2117 & 0.1227 & 0.5556 & 0.4810 & 0.4574 \\
    Actor & 0.2217 & 0.2235 & 0.2150 & 0.2154 & 0.2167 & 0.2220 & 0.2085 & 0.2128 \\
    Chameleon & 0.2272 & 0.2160 & 0.1906 & 0.1910 & 0.2299 & 0.2062 & 0.6528 & 0.6498 \\
    Squirrel & 0.2213 & 0.2150 & 0.1958 & 0.1955 & 0.2221 & 0.1925 & 0.5405 & 0.5441 \\
    Cora & 0.1222 & 0.1517 & 0.1525 & 0.1521 & 0.8100 & 0.3981 & 0.3427 & 0.3481 \\
    CiteSeer & 0.1529 & 0.1748 & 0.1751 & 0.1743 & 0.7355 & 0.1259 & 0.2951 & 0.3095 \\
    USA & 0.2366 & 0.2678 & 0.2517 & 0.2385 & 0.6978 & 0.2492 & 0.5022 & 0.5436 \\
    Europe & 0.2748 & 0.2794 & 0.2684 & 0.2596 & 0.4046 & 0.2460 & 0.4487 & 0.4742 \\
    Brazil & 0.2672 & 0.2941 & 0.2453 & 0.2110 & 0.4307 & 0.2326 & 0.5387 & 0.6038 \\
    \bottomrule
    \end{tabular}
    \caption{Total variation $TV(\bbY)$ (cf.~\eqref{eq:tv}) and edge homophily $h_{\rm edge}$ (cf.~\eqref{eq:edge_hom}) based on the node labels $\bby$ in multiple graph datasets measured across four different graphs: (i) the original graph $\ccalG$ and three $k$-NN graphs based on (ii) the original node features for $\ccalG^{\rm nn}_{k,\text{feat}}$, (iii) role-based structural attributes for $\ccalG^{\rm nn}_{k,\text{role}}$ or (iv) global structural attributes for $\ccalG^{\rm nn}_{k,\text{glob}}$.}
    \label{tab:smoothness}
\end{table*}

\section{The Role of Smoothness for GNNs}
\label{S:smoothness}

A crucial observation of GNNs is that most of the architectures in the literature have an inherent low-pass filtering behavior~\cite{nt2020revisiting}.
From a graph signal processing perspective, this means that high frequency components are attenuated, resulting in smoother node representations at the output layers~\cite{gama2020stability}.
To analyze this effect, we leverage the connection between signal smoothness (low $TV$) and its concentration in low-frequency spectral components.
GNN operations, like those in GCNs, can be seen as repeated applications of a normalized adjacency matrix $\hbA = \hbD^{-1/2}(\bbA + \bbI)\hbD^{-1/2}$. We therefore assess the TV of an $L$-layer GCN's filtering effect, approximated as $\hbA^L \bbx$ acting on an input signal $\bbx$.
Let the eigendecomposition of the graph Laplacian be $\hbL = \bbI - \hbA = \hbV\diag(\hblambda)\hbV^\top$ (assuming $\hbL$ is symmetric). The graph Fourier transform of $\bbx$ is $\tbx = \hbV^\top \bbx$. The repeated application of $\hbA$ in a GNN layer means the $L$-layer output's frequency representation is $\diag(\bbone-\hblambda)^L \tbx$.
Because the adjacency matrix $\hbA$ is normalized and includes self-loops, the eigenvalues of the Laplacian $\hblambda$ are approximately within the interval $[0,1.5]$~\cite{nt2020revisiting,wu2019simplifying}.
Thus, for the TV of the filtered signal with respect to the augmented adjacency matrix $\hbA$, we have that
\alna{
    TV(\hbA^L \bbx)^2
    &~=~&
    \| (\bbI-\hbA)\hbA^L\bbx \|_1^2
&\nonumber\\&
    &~\leq~&
    N
    \| \hbL(\bbI-\hbL)^L\bbx \|_2^2
&\nonumber\\&
    &~=~&
    N
    \| \diag(\hblambda) \diag( \bbone-\hblambda )^L \tbx \|_2^2
&\nonumber\\&
    &~=~&
    N
    \sum_{i=1}^N \hat{\lambda}_i^2 (1-\hat{\lambda}_i)^{2L} \tilde{x}_i^2,
\label{eq:tv_filt}}
so while $\hat{\lambda}_i \approx 0$ yields smaller terms of the sum in~\eqref{eq:tv_filt} because of the term $\hat{\lambda}_i^2$, so do the eigenvalues $\hat{\lambda}_i > 0.5$, particularly as $L$ increases, since this scales the $i$-th term of the sum in~\eqref{eq:tv_filt} by $(1-\hat{\lambda}_i)^{2L} \lesssim 0.5^{2L}$.
The core idea is that repeated application of a low-pass filter makes the output smoother (lower TV). Thus, GNNs inherently favor smoother signals, making them well-suited for graphs where labels are homophilic and thus already smooth.
Also, while in the analysis we focused on undirected graphs, a similar intuition holds for directed graphs by considering $\hbV^{-1}$ in lieu of $\hbV^\top$ when computing the frequency representation $\tbx$.

Based on the preceding analysis, GNN performance is expected to improve if the input graph has minimal ``false positive'' edges, i.e., edges connecting nodes from different classes. Ideally, this means $TV(\bbY)\approx 0$, $h_{\rm edge} \approx 1$, and $h_{\rm node}(i) \approx 1$ for all $i \in \ccalV$.
To formalize this intuition, we consider how well GNN performance can be improved by altering the original graph.
A feasible semi-supervised node classification task requires that the training labels $\bby_{\rm train}$ contain at least one node from each of the $C$ classes.
Furthermore, from a prediction point of view, it is clear that if all nodes belonging the same class end up having the same latent representation, then the true labels can be perfectly found from that latent representation. This simple idea is formalized in the following definition, with $P$ denoting the dimension of the latent space.

\begin{definition}\label{def:recover}
    Consider a node classification setup with $N$ nodes and $C$ classes. Let $\bby \in \{ 1,2,\dots, C \}^N$ be the vector collecting all the node labels and $\bbu^{(c)}\in\reals^{P}$ be a $P$-dimensional vector representing the $c$-th class. Then, we say that the $N$-dimensional label vector $\bby$ is recoverable from the $N\times P$ matrix $\bbU$ if the two following conditions hold: a) the $i$-th row of $\bbU_{i,:}$ is $\bbu^{(y_i)}$ for all $i$, and b) $\bbu^{(c)}\neq \bbu^{(c')}$ for all $c\neq c'$.
\end{definition}

\noindent
An illustrative example of the matrix $\bbU$ is the one-hot encoding $\bbY$ of node labels $\bby$, in which case we have $P = C$. 
For a second meaningful example, consider a matrix $\bbU$ that satisfies Definition~\ref{def:recover} with $P = C$, and suppose further that, for each vector $\bbu^{(c)}$, its largest entry corresponds precisely to the $c$-th position. 
Under these conditions, if the output of our GNN model $\Phi(\bbX;\bbA, \bbTheta)$ matches such a structure, applying a softmax operation will result in perfectly predicted labels. 
In practical scenarios, however, the GNN output typically will not exactly satisfy the conditions outlined in Definition~\ref{def:recover}. 
Nevertheless, one can evaluate prediction performance by examining how closely the GNN output $\hbZ$ approximates an ideal output $\bbZ^*$ that strictly fulfills Definition~\ref{def:recover}. 
This discrepancy depends inherently on the quality of the underlying graph structure $\bbA$ and the informative value of the node features $\bbX$.

Definition~\ref{def:recover} will be used in Theorem~ \ref{thm:error_bound} to characterize the performance of a GNN. 
Before stating the theorem, we need to introduce some additional concepts. First, we consider how false positive edges hinder the capacity of GNNs to recover $\bby$.
Let the observed graph $\ccalG$ have adjacency matrix $\bbA$. Consider a new graph $\ccalG^*$ without false positive edges, that is, with an adjacency matrix $\bbA^*$ such that 
\alna{
    A_{ij}^*
    &~:=~&
    \begin{cases}
        A_{ij}, & \text{if } y_i=y_j \\
        0, & \text{if } y_i \neq y_j
    \end{cases}
\label{eq:ideal_A}}
for all $i,j\in\ccalV$.
The graph $\ccalG^*$ is, by construction, perfectly homophilic with respect to the labels $\bby$. The false positive edges are then captured by the difference matrix $\bbDelta = \bbA - \bbA^*$. Intuitively, a sparser $\bbDelta$ (i.e., fewer false positive edges in the original graph) should increase the likelihood that the GNN $\Phi$ successfully recovers $\bby$.

In addition to a high-quality graph structure, the feasibility of recovering labels $\bby$ also hinges on how informative the node features $\bbX$ are.
Specifically, let $\bbx^{(c)} = (\bbY_{:,c}^\top\bbX) / (\bbY_{:,c}^\top\bbone)$ denote the mean feature vector for class $c$, where $\bbY_{:,c}$ is the column of the one-hot label matrix $\bbY$ corresponding to class $c$. We then define an idealized feature matrix $\bbX^* \in \reals^{N \times M}$ such that its $i$-th row is the mean feature vector of node $i$'s true class $y_i$
\alna{
    \bbX_{i,:}^* := \bbx^{(y_i)}
    =
    \frac{ \bbY_{:,y_i}^\top \bbX }{ \bbY_{:,y_i}^\top \bbone }
    \quad \forall~i\in\ccalV,
\label{eq:ideal_X}}
where $\bbX_{i,:}$ denotes the $i$-th row of $\bbX$. 
Thus, all nodes belonging to the same class $c$ share the identical row vector $\bbx^{(c)}$ in $\bbX^*$. 
For $\bbX$ to be sufficiently informative, we require that the vectors $\{ \bbx^{(c)} \}_{c=1}^C$ are distinct for each class, i.e., $\bbX^*$ recovers $\bby$ as in Definition~\ref{def:recover}.
This is analogous to Assumption 1 of~\cite{nt2020revisiting}, which was empirically verified in~\cite[Fig.~1]{nt2020revisiting}.
Essentially, this condition implies that the original node features $\bbX$ contain enough class-distinguishing information such that, if a GNN can effectively denoise $\bbX$ (e.g., by mapping noisy inputs towards these class means), it should be able to recover $\bby$.
Therefore, with few erroneous edges in the graph (a sparse $\bbDelta$) and sufficiently informative features (distinct class means in $\bbX$), our ability to recover the node labels $\bby$ should improve.
The subsequent theorem provides a formal basis for this intuition.

\begin{theorem} \label{thm:error_bound}
    Consider a two-layer GNN $\Phi:\reals^{N\times M}\rightarrow \reals^{N\times M'}$ predicting node labels $\bby \in \{1,\dots,C\}^{N}$,
    \alna{
        \Phi(\bbX;\bbA,\bbTheta)
        &~=~&
        \sigma_2\left(
        \hArw
        \sigma_1\left(
        \hArw \bbX \bbTheta^{(1)}
        \right) \bbTheta^{(2)}
        \right),
    \label{eq:gcn_thm}}
    with nonexpansive nonlinearities $\sigma_1,\sigma_2$ and learnable weights $\bbTheta = \{ \bbTheta^{(1)},\bbTheta^{(2)} \}$.
    Let $\hbZ := \Phi(\bbX; \bbA, \bbTheta)$ and $\bbZ^* := \Phi(\bbX^*;\bbA^*,\bbTheta)$ for $\bbA^*$ in~\eqref{eq:ideal_A} and $\bbX^*$ in~\eqref{eq:ideal_X}.
    Assume that $\bbX^*$ recovers $\bby$ and $\max_i \| [\bbX^* - \bbX]_{i,:} \|_2 \leq \alpha$ for some $\alpha \geq 0$.
    Then, $\bbZ^*$ recovers $\bby$, and the error between the predictions $\hbZ$ and $\bbZ^*$ is bounded above by
    \alna{
        \| \bbZ^* - \hbZ \|_F
        &\,\leq\,&
        \rho_1 \rho_2 \!
        \left(
            \alpha \sqrt{N}
            +
            2(1 + \sqrt{N})
            \| \bbDelta \|_F
            \|\bbX\|_F
        \right)
    \label{eq:err_bnd}}
    for $\rho_1=\|\bbTheta^{(1)}\|_2$, $\rho_2=\|\bbTheta^{(2)}\|_2$, and $\bbDelta = \bbA - \bbA^*$.
\end{theorem}

\begin{figure*}[t!]
    \centering
    \includegraphics[width=\textwidth]{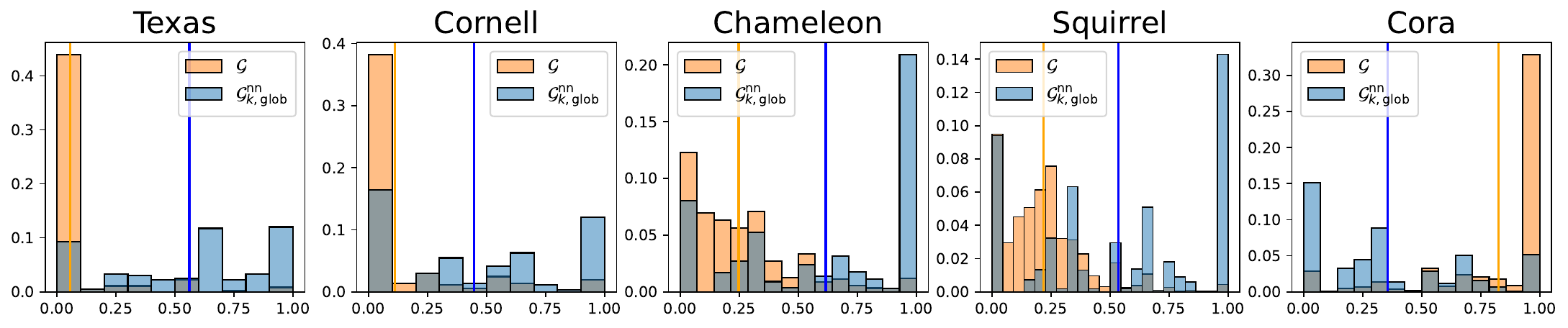}
    \caption{Histograms of node homophily measurements $h_{\rm node}$ (cf.~\eqref{eq:node_hom}) based on the node labels $\bby$ in multiple graph datasets, measured using the original graph $\ccalG$ and a $k$-NN graph using global structural attributes $\ccalG^{\rm nn}_{k,{\rm glob}}$. 
    The $x$-axis corresponds to homophily level $h_{\rm node}$, while the $y$-axis represents the proportion of nodes in each bin.}
    \label{fig:homophilies}
\end{figure*}

\noindent
The proof of Theorem~\ref{thm:error_bound} is provided in Appendix~\ref{app:thm_proof}.
The inequality in~\eqref{eq:err_bnd} bounds how different the GNN node embeddings $\hbZ$ are from the ideal ones $\bbZ^*$, which map each node to a distinct point that determines its class label.
As $\| \bbZ^* - \hbZ \|_F$ decreases, the embeddings in $\hbZ$ move closer to the distinct class-wise points in $\bbZ^*$ and hence become increasingly distinguishable across classes.
This bound highlights two primary sources of deviation from the ideal prediction $\bbZ^*$. The first term, scaled by $\alpha\sqrt{N}$, quantifies the impact of noise in the input node features $\bbX$.
For a learning problem to be tractable, this feature noise $\alpha$ is generally assumed to be sufficiently small. The second term, involving $\|\bbDelta\|_F$, directly reflects the influence of false positive edges in the graph $\bbA$.
This term underscores the homophily assumption inherent in GNNs: edges are expected to connect similar nodes, and deviations (false positives) contribute to prediction error.
The scaling $\rho_1 \rho_2$ reflects the magnitudes of the GNN weight matrices $\bbTheta^{(1)}$ and $\bbTheta^{(2)}$, which can be bounded during training via common techniques such as $\ell_2$ regularization.
Therefore, Theorem~\ref{thm:error_bound} demonstrates that using graphs on which node labels are more homophilic by eliminating false positive edges can improve the ability of GNNs to accurately recover unseen labels.

\section{Structure-Based Neighbor Discovery}
\label{S:neighbor_discovery}

\begin{figure}
    \centering
    \includegraphics[width=\linewidth]{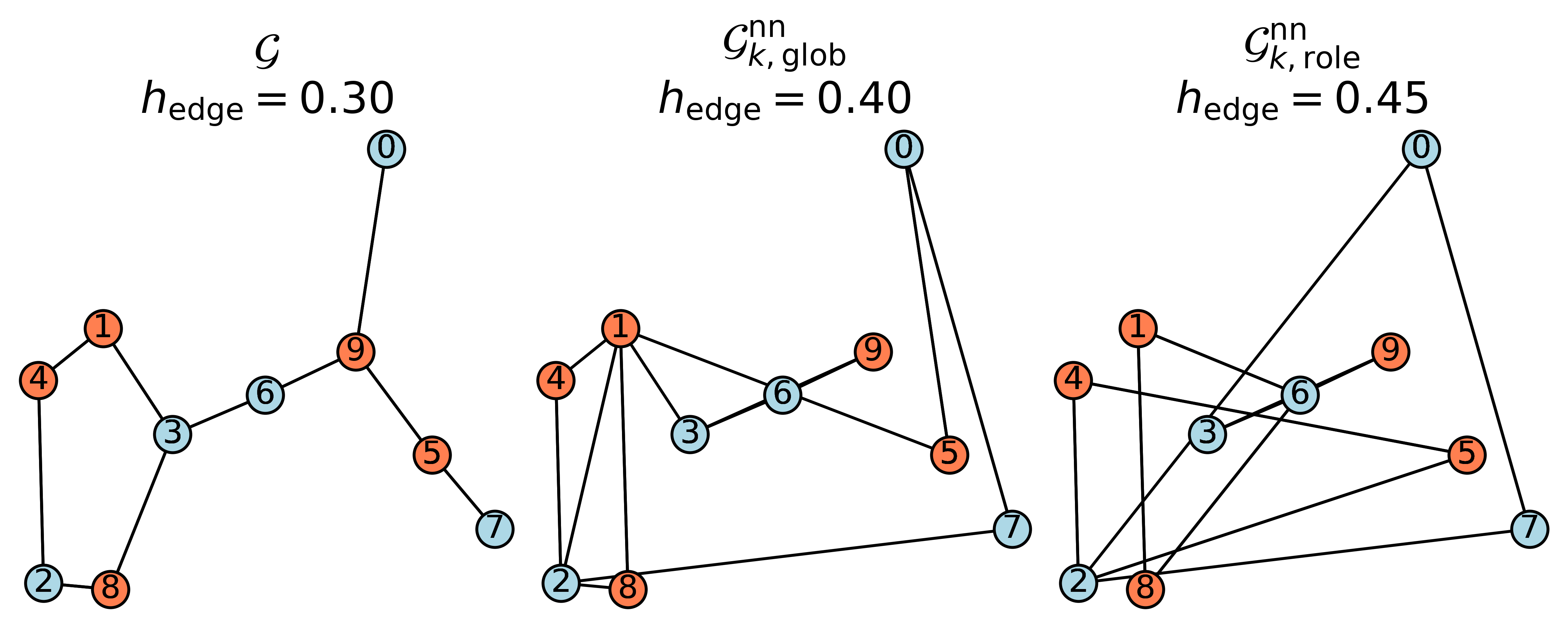}
    \caption{Structure-based neighbor discovery on a toy graph.
    (Left) The original graph $\mathcal{G}$ exhibits low edge homophily.
    (Center) A $k$-NN graph $\ccalG^{\rm nn}_{k, {\rm glob}}$ constructed using global structural attributes.
    (Right) A $k$-NN graph $\ccalG^{\rm nn}_{k, {\rm role}}$ constructed using role-based structural attributes.
    Both alternative graphs, $\ccalG^{\rm nn}_{k, {\rm glob}}$ and $\ccalG^{\rm nn}_{k, {\rm role}}$, achieve higher edge homophily by increased connections between nodes of the same class (indicated by color) compared to the original graph.}
    \label{fig:barbell_graph}
\end{figure}

Motivated by Theorem~\ref{thm:error_bound}, which underscores the benefit of graphs with high label homophily, we propose using structural attributes derived from the original graph $\ccalG$ to construct new graphs where such homophily is enhanced. While various information sources could be used, we exemplify our approach using three types of node attributes:

\begin{enumerate}[align=parleft, leftmargin=*]
    \item \textbf{Role-based features:} capture the local connectivity patterns of nodes, such as degree and triangle counts, reflecting its structural role in its immediate surroundings~\cite{henderson2012rolx, guo20role}.

    \item \textbf{Global features:} describe the importance or position relative of a node to the entire graph, often via centrality measures~\cite{newman2018networks}.

    \item \textbf{Node features:}
    the given node features $\bbX$.
    A common strategy is to build a new graph based on the similarity of these features~\cite{jin2021similarity,li2024seeking}.
\end{enumerate}

Using original node features $\bbX$ to construct a graph can be effective if they are highly informative for predicting labels. However, this approach discards the original graph structure entirely and can propagate errors if $\bbX$ is noisy or imperfect~\cite{jin2021similarity}.
Alternatively, our focus is on leveraging well-established, interpretable, and computationally efficient graph-theoretic metrics for role-based and global structural attributes (detailed in Appendix~\ref{app:struc_features}). As we will numerically illustrate later in, e.g., Table~\ref{tab:smoothness}, graphs constructed from these chosen structural attributes often exhibit significantly higher label homophily on real-world datasets compared to the original graph. While learned node embeddings may also serve as features $\{\bbf_i\}_{i=1}^N$~\cite{suresh2021breaking}, they are harder to interpret.
Moreover, as we show in our numerical evaluation, they do not consistently outperform well-chosen graph-theoretic features.
Motivated by this, next we detail how to compute new graphs from these attribute types.

Let $\bbf_i \in \reals^F$ denote the $F$-dimensional feature vector associated with node $i\in\ccalV$. This vector can represent role-based attributes, global attributes, or the original node features $\bbX_{i,:}$, in which case $F=M$.
We construct alternative graphs by measuring the similarity between these vectors $\{\bbf_i\}_{i=1}^N$.
Specifically, we compute a distance matrix $\bbB \in\reals^{N\times N}$, where $B_{ij} = \|\bbf_i-\bbf_j\|_2^2$ is the squared Euclidean distance between the attribute vectors of nodes $i$ and $j$.

We consider two kinds of graphs: $k$-NN graphs, denoted $\ccalG^{\rm nn}_k$, and $\epsilon$-ball graphs, denoted $\ccalG^{\rm ball}_{\epsilon}$. 
For the former, let $\ccalD_k(i) \subset \ccalV$ denote the set of $k$ nodes in $\ccalV$ that have the smallest Euclidean distances to node $i$ based on $\bbB$. 
An undirected edge $(i,j)$ exists in $\ccalG^{\rm nn}_k$ if $i\in\ccalD_k(j)$ or $j\in\ccalD_k(i)$, that is, the new edge set for $\ccalG^{\rm nn}_k$ is
\alna{
    \ccalE^{\rm nn}_{k}
    &~=~&
    \{
        (i,j)
        ~|~
        i \in \ccalD_k(j)
        {\rm~or~}
        j\in\ccalD_k(i), ~
        i,j\in\ccalV
    \}.
\nonumber}
For $\ccalG^{\rm ball}_{\epsilon}$, an edge $(i,j)$ exists if and only if the distance $B_{ij}$ is less than a predefined threshold $\epsilon$, resulting in the edge set
\alna{
    \ccalE^{\rm ball}_{\epsilon}
    &~=~&
    \{
        (i,j)
        ~|~
        B_{ij} < \epsilon, ~
        i,j\in\ccalV
    \}.
\nonumber}
These graph construction methods are straightforward, computationally manageable, and directly interpretable when based on well-understood structural attributes, like the role-based and global features.
For a visual example, Fig.~\ref{fig:barbell_graph} depicts how new graphs derived from role-based and global structural attributes on a toy graph exhibit enhanced edge homophily compared to the original.

Crucially, graphs constructed from these structural attributes often prove more informative for GNN-based node classification tasks, particularly by enhancing label homophily on datasets where the original graph is heterophilic.
This is illustrated in Table~\ref{tab:smoothness}, which presents the total variation $TV(\bbY)$ and edge homophily $h_{\rm edge}$ of node labels across diverse datasets.
We compare the original graph $\ccalG$ with $k=3$ $k$-NN graphs constructed using: (i) original node features $\bbX$, denoted as $\ccalG^{\rm nn}_{k,{\rm feat}}$; (ii) our proposed role-based structural attributes, denoted as $\ccalG^{\rm nn}_{k,{\rm role}}$; and (iii) our proposed global structural attributes, denoted as $\ccalG^{\rm nn}_{k, {\rm glob}}$.
For many heterophilic datasets~\cite{pei2020geom} (e.g., Texas, Wisconsin, Chameleon, Squirrel), Table~\ref{tab:smoothness} shows that $TV(\bbY)$ is generally lower and $h_{\rm edge}$ is higher for these alternative $k$-NN graphs compared to the original $\ccalG$.
This indicates that node labels exhibit greater smoothness on these newly constructed graphs.
Conversely, for inherently homophilic datasets like Cora~\cite{mccallum2000cora} and CiteSeer~\cite{giles98citeseer}, the original graph already possesses high $h_{\rm edge}$ and low $TV(\bbY)$, and our $k$-NN constructions typically result in less homophilic graphs by these metrics. 
Furthermore, Fig.~\ref{fig:homophilies} displays the node homophily $h_{\rm node}$ for four datasets and aligns with our previous discussion: for heterophilic datasets, the $k$-NN graphs $\ccalG^{\rm nn}_{k,{\rm glob}}$ exhibit node homophily distributions that are more concentrated toward higher $h_{\rm node}$ values than those of the original graphs.
This suggests that for datasets where traditional GNNs struggle due to low original homophily, our structure-based graphs offer more suitable topologies.

\begin{table}[t]
    \centering
    \begin{tabular}{l|cccc}
    \toprule
    $\| \bbDelta \|_F / N$ & $\ccalG$ & $\ccalG^{\rm nn}_{k,{\rm feat}}$ & $\ccalG^{\rm nn}_{k,{\rm role}}$ & $\ccalG^{\rm nn}_{k,{\rm glob}}$ \\ \midrule
    Texas & 0.0931 & 0.0848 & 0.0792 & 0.0829 \\
    Wisconsin & 0.0811 & 0.0705 & 0.0813 & 0.0814 \\
    Cornell & 0.0879 & 0.0850 & 0.0923 & 0.0945 \\
    Actor & 0.0202 & 0.0175 & 0.0177 & 0.0176 \\
    Chameleon & 0.0730 & 0.0322 & 0.0216 & 0.0217 \\
    Squirrel & 0.0789 & 0.0216 & 0.0158 & 0.0159 \\
    Cora & 0.0165 & 0.0256 & 0.0267 & 0.0265 \\
    CiteSeer & 0.0147 & 0.0277 & 0.0248 & 0.0245 \\
    USA & 0.0539 & 0.0435 & 0.0352 & 0.0335 \\
    Europe & 0.1497 & 0.0752 & 0.0643 & 0.0625 \\
    Brazil & 0.1824 & 0.1316 & 0.1013 & 0.0960 \\
    \bottomrule
    \end{tabular}
    \caption{Normalized count of false positive edges, $\| \bbDelta \|_F / N$, for the original graph $\ccalG$ and $k$-NN graphs constructed from node features $\ccalG^{\rm nn}_{k,{\rm feat}}$, role-based attributes $\ccalG^{\rm nn}_{k,{\rm role}}$, and global attributes $\ccalG^{\rm nn}_{k,{\rm glob}}$ across various datasets.}
    \label{tab:false_pos_count}
\end{table}

Indeed, graphs with higher label homophily and consequently fewer false positive edges are theoretically expected to improve GNN performance, as suggested by Theorem~\ref{thm:error_bound}, where the error bound~\eqref{eq:err_bnd} depends on $\|\bbDelta\|_F$.
Table~\ref{tab:false_pos_count} quantifies this normalized term $\|\bbDelta\|_F/N$.
Consistent with the homophily metrics in Table~\ref{tab:smoothness}, for inherently homophilic datasets like Cora and CiteSeer, the original graph $\ccalG$ has a lower $\|\bbDelta\|_F/N$ than the $k$-NN alternatives. However, for most other datasets, particularly the heterophilic ones, our $k$-NN graphs constructed from structural attributes (role-based or global) and even from original features often exhibit a lower count of false positive edges than the original graph.
This aligns with the theoretical benefits suggested by Theorem~\ref{thm:error_bound} and further supports their use as effective alternatives for GNN-based node classification in such settings.

\begin{figure}[b]
    \centering
    \includegraphics[width=\linewidth]{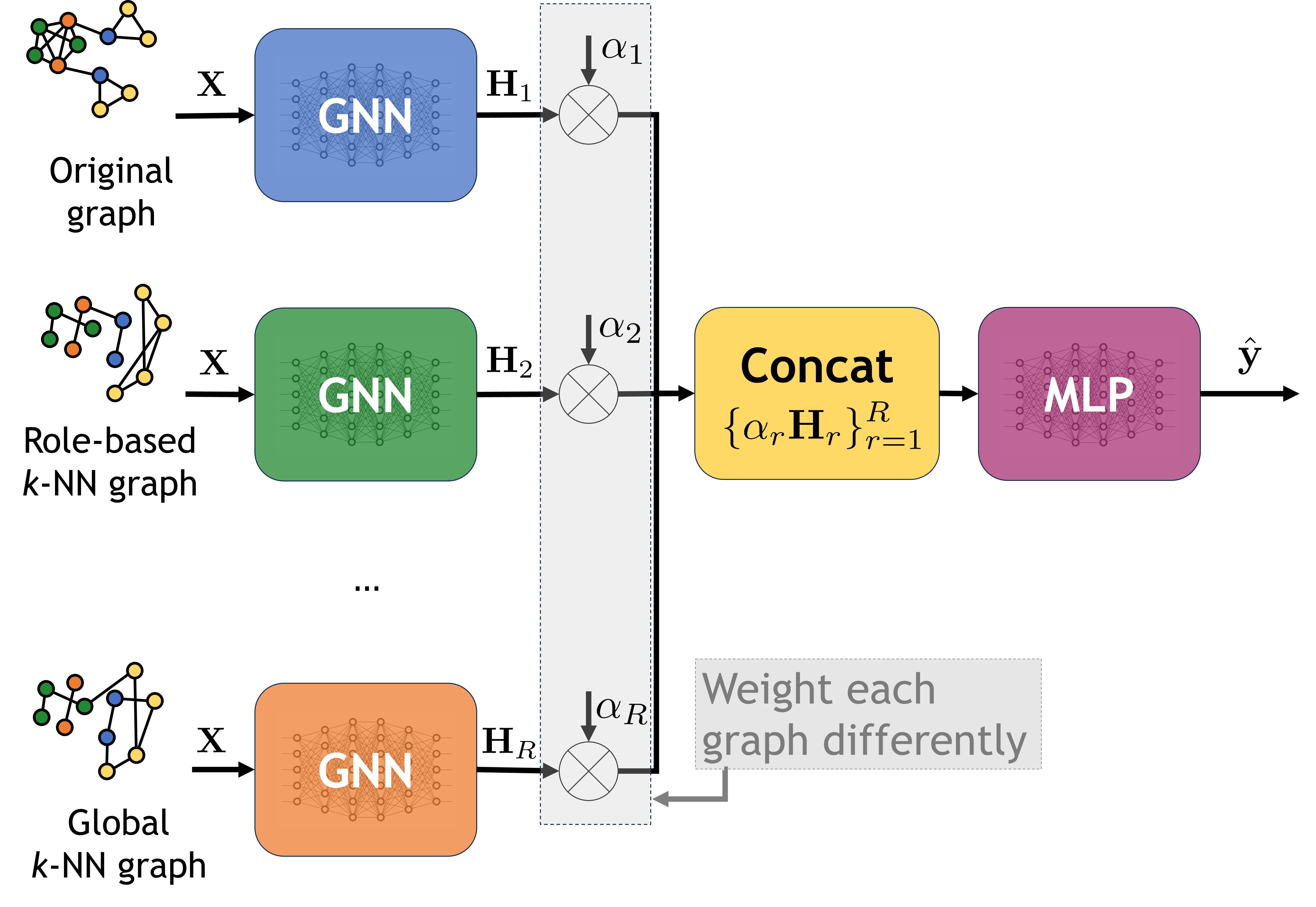}
    \caption{Proposed architecture exemplified using three input graphs.
    We assign each graph to a GNN layer with adaptive weights denoting the importance of each graph.
    The resultant graph representations are combined via concatenation and fed to a fully connected network for the final representation.}
    \label{fig:architecture}
\end{figure}

\section{Structure-Guided GNN}
\label{S:sg-gnn}

While individual structurally-informed graphs can enhance homophily, as we empirically demonstrate in Section~\ref{S:exp}, the optimal choice of a single alternative graph is often unknown a priori and may vary across datasets or tasks. A natural extension is therefore to leverage multiple graph views concurrently, exploiting diverse types of connections.
To this end, we propose a novel architecture termed Structure-Guided GNN (SG-GNN) that simultaneously integrates several graphs defined on the same node set but with different edge structures, analogous to tensor-based GNNs~\cite{ioannidis2020tensor}.

Specifically, we consider a set of $R$ graphs, comprising the original graph $\ccalG$ and $R-1$ new graphs derived from structural attributes (e.g., role-based and global features), as well as from different node embeddings.
Then, SG-GNN exploits the $R$ graphs to process the input features $\bbX$ in parallel. As depicted in Fig.~\ref{fig:architecture}, each graph $\ccalG_r$ with its corresponding adjacency matrix $\bbA_r$ is fed into a different branch within the GNN. For the $r$-th graph, this operation, using the initial node features $\bbX$, produces an output representation
\alna{
    \bbH_r
    &~=~&
    \phi( \bbX; \bbA_r, \bbTheta_r ),
\label{eq:r_gcn_layer}}
where $\phi$ is a GNN layer function [cf.~\eqref{eq:gcn_layer}] with parameters $\bbTheta_r$ specific to graph $\ccalG_r$.
The outputs $\{\bbH_r\}_{r=1}^R$ from these parallel branches are then aggregated using adaptively learned 
weights $\{\alpha_{r}\}_{r=1}^R$. 
After scaling each $\bbH_r$ by the corresponding weight $\alpha_r$, we concatenate these scaled representations and feed them to a Multi-Layer Perceptron (MLP), denoted as $\psi$, obtaining the final node representations $\hby$ as follows
\alna{
    \hby
    &~=~&
    \psi\left( {\rm concat} \left( \{ \alpha_r \bbH_r \}_{r=1}^R \right) \right),
\nonumber}
where $\text{concat}(\cdot)$ denotes horizontal concatenation. The weights $\alpha_r$ are constrained such that $\sum_{r=1}^R \alpha_r = 1$ and $\alpha_r \ge 0$ via a softmax function over learnable pre-activation scores.
The learned coefficients $\{\alpha_{r}\}_{r=1}^R$ represent the relative importance of each graph for the task of interest, where a larger $\alpha_r$ indicates that the $r$-th graph is more structurally informative.
This adaptive weighting not only helps the SG-GNN compete with models that use only a single graph but also offers interpretability by highlighting which graph views are most crucial.

The rationale underlying SG-GNN is that harnessing information from multiple graph views can lead to a better performance than relying on a single, potentially imperfect, graph. We can formalize this by considering the likelihood of finding a graph with high label homophily (i.e., few false positive edges).
For any of the $R$ candidate graphs to be beneficial, it should ideally minimize false positive connections. We model this by assuming that for any given pair of nodes with different labels, the probability of an erroneous edge existing between them in any single graph $\ccalG_r$ is at most $q$. Hence, a small $q$ implies that the features used to construct $\ccalG_r$ (be they original features $\bbX$ or our structural attributes) are semantically informative enough to reduce spurious connections between dissimilar nodes.
The following proposition proves that considering multiple such graphs increases the probability of finding at least one graph that is perfectly homophilic (i.e., has no false positive edges). For simplicity and without loss of generality, we assume classes are equally sized, with $K := N/C$ nodes per class.

\begin{proposition}\label{prop:sbm}
    Let $\{ \ccalG_r \}_{r=1}^R = \{ (\ccalV,\ccalE_r) \}_{r=1}^R$ be a set of independent graphs with shared node labels $\bby \in \{1,\dots,C\}^N$.
    For any graph $\ccalG_r$ and any pair of nodes $(i,j)$ with $y_i \neq y_j$, assume the probability of an edge $(i,j)$ existing in $\ccalE_r$ is $\mbP[(i,j)\in\ccalE_r | y_i\neq y_j] \leq q$, where edge likelihoods are independent across all node pairs.
    The probability that at least one graph $\ccalG_r$ has no false positive edges is bounded below by
    \alna{
        \mbP\left[ \bigcup_{r=1}^R \left\{
            y_i=y_j~\forall \, (i,j) \in \ccalE_r
        \right\} \right]
    &\nonumber\\&
    \qquad\qquad\qquad\quad
    \geq
    1 - 
    \Big(
        1- (1-q)^{\frac{N^2(C-1)}{2C}}
    \Big)^{\!R}.
    \label{eq:sbm_prob}}
\end{proposition}

\begin{table}[t]
    \centering
    \begin{tabular}{l|cccc}
    \toprule
    $\hat{q}$ & $\mathcal{G}$ & $\ccalG^{\rm nn}_{k,{\rm feat}}$ & $\ccalG^{\rm nn}_{k,{\rm role}}$ & $\ccalG^{\rm nn}_{k,{\rm glob}}$ \\
    \midrule
    Texas & 0.0150 & 0.0129 & 0.0073 & 0.0108 \\
    Wisconsin & 0.0095 & 0.0064 & 0.0078 & 0.0094 \\
    Cornell & 0.0121 & 0.0085 & 0.0085 & 0.0101 \\
    Actor & 0.0005 & 0.0002 & 0.0003 & 0.0004 \\
    Chameleon & 0.0086 & 0.0005 & 0.0006 & 0.0005 \\
    Squirrel & 0.0113 & 0.0003 & 0.0003 & 0.0003 \\
    Cora & 0.0003 & 0.0006 & 0.0008 & 0.0008 \\
    CiteSeer & 0.0003 & 0.0014 & 0.0005 & 0.0005 \\
    USA & 0.0032 & 0.0000 & 0.0014 & 0.0015 \\
    Europe & 0.0000 & 0.0001 & 0.0056 & 0.0053 \\
    Brazil & 0.0527 & 0.0000 & 0.0120 & 0.0117 \\
    \bottomrule
    \end{tabular}
    \caption{Empirical estimates of the false positive edge probability, $\hat{q}$, for the original graph and various $k$-NN graphs across datasets.}
    \label{tab:q_values}
\end{table}

\noindent
The proof of Proposition~\ref{prop:sbm} can be found in Appendix~\ref{app:prop1_proof}.
The key implication of~\eqref{eq:sbm_prob} is that, as the number of considered graphs $R$ increases, the term $(1- (1-q)^{\frac{N^2(C-1)}{2C}})^R$ decreases. Consequently, the lower bound on the probability of finding at least one perfectly homophilic graph increases with $R$.
While the graphs derived from the same original graph $\ccalG$ in our approach may not satisfy all the independence assumptions, this theoretical result still provides valuable intuition: incorporating more diverse graph views, each constructed to capture different structural aspects or feature similarities, is likely to improve the chances of identifying at least one structure (or a beneficial combination thereof) that is more homophilic than the original, thereby motivating the multi-graph approach of SG-GNN.
In practice, however, the decision on the number of views $R$ involves a trade-off: while the theoretical bound improves with $R$, the actual gains may saturate due to statistical dependencies among the constructed graphs as highly correlated graphs may not contribute new information, in addition to the increasing computational and memory costs.

\subsection{Node-specific aggregation}
\label{Ss:node-specific}

To further enhance model expressiveness and provide finer-grained interpretability, we extend the SG-GNN to incorporate node-specific aggregation weights. Instead of global weights $\{\alpha_r\}_{r=1}^R$, each node $i \in \ccalV$ now learns its own set of coefficients $\{ \alpha_{ir} \}_{r=1}^R$ to weigh the contributions from the $R$ different graph views. This allows each node to dynamically emphasize graph structures that are most relevant to its local context or classification, akin to node-level attention mechanisms over graphs~\cite{velickovic2018graph}. This approach can be particularly beneficial as the optimal graph structure for ensuring homophilic connections may vary from one node to another.
A graph that is highly homophilic for node $i$ might not be for node $j$, as observed in the high variance of the histograms shown in Fig.~\ref{fig:homophilies}.

This node-level adaptability highlights the value of considering multiple graphs. We can formalize this intuition similarly to Proposition~\ref{prop:sbm}. The following proposition shows that increasing the number of graph views improves the likelihood that \emph{every node} finds at least one graph where its local neighborhood is purely homophilic.

\begin{proposition}\label{prop:sbm_node}
    Let $\{ \ccalG_r \}_{r=1}^R = \{ (\ccalV,\ccalE_r) \}_{r=1}^R$ be a set of independent graphs with shared node labels $\bby \in \{1,\dots,C\}^N$.
    For any graph $\ccalG_r$ and any pair of nodes $(i,j)$ with $y_i \neq y_j$, assume the probability of an edge $(i,j)$ existing in $\ccalE_r$ is $\mbP[(i,j)\in\ccalE_r | y_i\neq y_j] \leq q$, where edge likelihoods are independent across all node pairs.
    Then, the probability that for every node $i\in\ccalV$, there exists at least one graph $\ccalG_r$ in which all neighbors $j \in \ccalN_r(i)$ have the same class as node $i$ (i.e., $y_j = y_i$), is bounded below by
    \alna{
        \mbP\left[
            \bigcap_{i=1}^N
            \bigcup_{r=1}^R \left\{ 
            y_i=y_j~\forall \, j\in\ccalN_r(i)
        \right\} \right]
        &\nonumber\\&
        \qquad\qquad\qquad\quad
        \geq
        \Big(1 - 
        \Big(1 - (1-q)^{ \frac{N(C-1)}{C} }
        \Big)^{\!R\,}
        \Big)^{\!N}.
    \label{eq:sbm_prob_node}}
\end{proposition}

\noindent
We prove Proposition~\ref{prop:sbm_node} in Appendix~\ref{app:prop2_proof}.
The lower bound in~\eqref{eq:sbm_prob_node} reveals distinct scaling behaviors compared to the global graph bound in~\eqref{eq:sbm_prob}. While both benefit from an increasing number of graph views $R$, their dependence on $N$ differs significantly. The exponent $\frac{N^2(C-1)}{2C}$ in~\eqref{eq:sbm_prob} reflects the challenge of ensuring homophily across all $\sim N^2$ potential inter-class pairs for a single graph to be globally perfect. In contrast, the exponent $\frac{N(C-1)}{C}$ in~\eqref{eq:sbm_prob_node} pertains to ensuring a homophilic neighborhood for one node, considering its potential connections to $\sim N$ other nodes. Although this less stringent per-node requirement is then raised to the power of $N$ (for the $\bigcap_{i=1}^N$ condition), the base probability of a single node finding a good local view is higher due to the linear (not squared) $N$ in its inner exponent. This highlights the advantage of node-specific adaptation: it is probabilistically easier to find a suitable local homophilic neighborhood for each node from $R$ views than to find one globally perfect graph. Thus, even if $N$ is large, increasing $R$ offers a practical path towards ensuring that most, if not all, nodes can leverage a locally beneficial graph structure.

The probabilistic assumptions in Propositions~\ref{prop:sbm} and~\ref{prop:sbm_node} can be connected to empirical observations.
Given a graph $\ccalG$, we can compute the empirical estimate of $q$, denoted as $\hat{q}$, as
\alna{
    \hat{q}(\ccalG)
    &~=~&
    \frac{ |(i,j) \in \ccalE ~:\, y_i\neq y_j | }{ |(i,j) \in \ccalV\times\ccalV ~:\, y_i\neq y_j | }.
\nonumber}
As shown in Table~\ref{tab:q_values}, the estimate $\hat{q}$ for the original graph is often higher than for our alternative graphs ($\ccalG^{\rm nn}_{k,\text{feat}}, \ccalG^{\rm nn}_{k,\text{role}}, \ccalG^{\rm nn}_{k,\text{glob}}$), especially on heterophilic datasets. These lower $\hat{q}$ values for our constructed graphs imply they are less likely to contain false positive edges (or more likely to have homophilic neighborhoods), aligning with the propositions' intuition that such graphs are beneficial and more readily found when considering multiple views or better construction methods.

\subsection{SG-GNN with Multiple Layers}
\label{Ss:full_sg-gnn}

The adaptive aggregation mechanism of SG-GNN can be extended by stacking multiple such layers to create a deeper architecture. In a multi-layer SG-GNN, each layer $\ell$ processes the hidden representations $\bbH^{(\ell)}$ from the previous layer (or input features $\bbX$ for the first layer) using the $R$ parallel graph-specific branches. The intermediate representation for the $r$-th graph at layer $\ell+1$ is thus computed as
\alna{
    \bbH_r^{(\ell+1)}
    &~=~&
    \phi( \bbH^{(\ell)}; \bbA_r, \bbTheta_r ).
\label{eq:rl_gcn_layer}}
The output of the $\ell$-th SG-GNN layer, $\bbH^{(\ell+1)}$, is then obtained by concatenating these intermediate graph-specific representations $\{\bbH_r^{(\ell+1)}\}_{r=1}^R$, followed by an MLP $\psi^{(\ell)}$
\alna{
    \bbH^{(\ell+1)}
    &~=~&
    \psi^{(\ell)}\left( {\rm concat} \left( \{ \bbH_r^{(\ell+1)} \}_{r=1}^R \right) \right).
\nonumber}
In this formulation, the MLP $\psi^{(\ell)}$ is responsible for learning how to best combine and transform the features derived from the different graph structures at each layer. The relative importance of each graph view is implicitly learned through the weights of the MLP as it processes the concatenated representations from $\bbH^{(\ell)} \in \reals^{N \times M_{\ell} }$ to produce $\bbH^{(\ell+1)} \in \reals^{N \times M_{\ell+1}}$.
This multi-layer architecture is not only more expressive but also more adaptive to information from the various graph structures.
The model may leverage different graphs at different depths, such as focusing on local role-based information in earlier layers and broader global structures in later ones, yielding increasingly rich representations~\cite{xu2018gnns}.

\section{Numerical Experiments}
\label{S:exp}

\begin{table*}[t]
    \centering
\begin{tabular}{c|cccccccc}
\toprule
 & \multicolumn{4}{c}{GCN} & \multicolumn{4}{c}{FBGNN} \\
 \cmidrule(lr){2-5} \cmidrule(lr){6-9}
 & Wisconsin & Cornell & Actor & Cora & Wisconsin & Cornell & Actor & Cora \\
\midrule
$\mathcal{G}$ & 44.23 $\pm$ 1.1 & 36.32 $\pm$ 1.45 & 28.63 $\pm$ 0.17 & \textbf{87.39 $\pm$ 0.22} & 81.35 $\pm$ 0.84 & 68.95 $\pm$ 0.94 & 33.76 $\pm$ 0.12 & \textbf{87.44 $\pm$ 0.22} \\ \midrule
$\ccalG^{\rm ball}_{\epsilon, {\rm feat}}$ & \textbf{81.54 $\pm$ 0.73} & \underline{61.58 $\pm$ 1.58} & \textbf{35.90 $\pm$ 0.16} & 72.3 $\pm$ 0.22 & 84.23 $\pm$ 0.81 & 72.89 $\pm$ 1.41 & 35.77 $\pm$ 0.11 & 75.94 $\pm$ 0.17 \\
$\ccalG^{\rm nn}_{k, {\rm feat}}$ & 72.88 $\pm$ 1.07 & \textbf{69.74 $\pm$ 1.08} & 30.3 $\pm$ 0.17 & 74.52 $\pm$ 0.17 & 79.23 $\pm$ 1.05 & 65.79 $\pm$ 1.1 & 35.91 $\pm$ 0.19 & 76.1 $\pm$ 0.2 \\
$\ccalG^{\rm ball}_{\epsilon, {\rm role}}$ & 75.96 $\pm$ 1.0 & 56.32 $\pm$ 1.33 & 34.32 $\pm$ 0.24 & 59.28 $\pm$ 0.27 & 84.23 $\pm$ 0.84 & \textbf{74.74 $\pm$ 0.68} & \textbf{36.11 $\pm$ 0.11} & 74.56 $\pm$ 0.18 \\
$\ccalG^{\rm nn}_{k, {\rm role}}$ & 60.0 $\pm$ 0.77 & 49.74 $\pm$ 1.11 & 29.48 $\pm$ 0.13 & 49.28 $\pm$ 0.3 & 83.46 $\pm$ 0.83 & 65.79 $\pm$ 0.82 & 34.23 $\pm$ 0.19 & 73.47 $\pm$ 0.16 \\
$\ccalG^{\rm ball}_{\epsilon, {\rm glob}}$ & 73.27 $\pm$ 0.76 & 55.79 $\pm$ 1.45 & \underline{35.21 $\pm$ 0.23} & 62.39 $\pm$ 0.24 & 82.5 $\pm$ 0.88 & 69.47 $\pm$ 1.02 & 35.9 $\pm$ 0.14 & 75.88 $\pm$ 0.17 \\
$\ccalG^{\rm nn}_{k, {\rm glob}}$ & 60.77 $\pm$ 1.2 & 48.68 $\pm$ 1.19 & 27.42 $\pm$ 0.18 & 48.97 $\pm$ 0.34 & 78.65 $\pm$ 0.81 & 67.11 $\pm$ 1.11 & 32.73 $\pm$ 0.15 & 73.86 $\pm$ 0.21 \\ \midrule
$\ccalG^{\rm ball}_{\epsilon, {\rm DW}}$ & 62.88 $\pm$ 1.16 & 57.11 $\pm$ 1.12 & 35.04 $\pm$ 0.26 & 85.96 $\pm$ 0.25 & 84.62 $\pm$ 0.76 & 68.68 $\pm$ 1.19 & 35.09 $\pm$ 0.12 & 85.83 $\pm$ 0.22 \\
$\ccalG^{\rm nn}_{k, {\rm DW}}$ & 48.46 $\pm$ 1.06 & 34.47 $\pm$ 1.12 & 29.0 $\pm$ 0.1 & 82.9 $\pm$ 0.21 & 80.77 $\pm$ 0.87 & 57.89 $\pm$ 1.94 & 33.77 $\pm$ 0.12 & 83.42 $\pm$ 0.2 \\
$\ccalG^{\rm ball}_{\epsilon, {\rm N2V}}$ & 67.69 $\pm$ 0.95 & 61.05 $\pm$ 1.3 & 35.19 $\pm$ 0.18 & \underline{86.51 $\pm$ 0.17} & \underline{85.38 $\pm$ 0.74} & 71.32 $\pm$ 0.98 & 35.24 $\pm$ 0.08 & \underline{86.49 $\pm$ 0.2} \\
$\ccalG^{\rm nn}_{k, {\rm N2V}}$ & 45.58 $\pm$ 0.6 & 34.47 $\pm$ 1.28 & 27.52 $\pm$ 0.11 & 83.9 $\pm$ 0.24 & 77.88 $\pm$ 0.83 & 65.53 $\pm$ 1.61 & 32.51 $\pm$ 0.11 & 83.64 $\pm$ 0.2 \\
$\ccalG^{\rm ball}_{\epsilon, {\rm S2V}}$ & \underline{76.73 $\pm$ 0.94} & 56.32 $\pm$ 1.39 & 34.36 $\pm$ 0.2 & 61.1 $\pm$ 0.19 & \textbf{86.15 $\pm$ 0.96} & \underline{74.47 $\pm$ 1.2} & \underline{35.93 $\pm$ 0.1} & 76.07 $\pm$ 0.18 \\
$\ccalG^{\rm nn}_{k, {\rm S2V}}$ & 55.38 $\pm$ 0.82 & 41.05 $\pm$ 1.46 & 28.07 $\pm$ 0.21 & 51.97 $\pm$ 0.17 & 80.19 $\pm$ 0.72 & 61.05 $\pm$ 1.79 & 33.96 $\pm$ 0.15 & 73.6 $\pm$ 0.24 \\
$\ccalG^{\rm nn}_{k, {\rm GW}}$ & 60.19 $\pm$ 1.27 & 52.11 $\pm$ 1.14 & 30.37 $\pm$ 0.15 & 49.25 $\pm$ 0.26 & 81.92 $\pm$ 0.89 & 66.05 $\pm$ 1.37 & 34.7 $\pm$ 0.13 & 74.67 $\pm$ 0.24 \\
\bottomrule
\end{tabular}
\caption{Node classification accuracy (\%) on various datasets using GCN and FBGNN layers. We compare the original graph $\mathcal{G}$ with $k$-NN graphs $\mathcal{G}_k$ and $\epsilon$-ball graphs $\mathcal{G}_\epsilon$ constructed using: original node features (feat), our proposed role-based (role) and global-based (global) structural attributes, and established node embedding methods, including DeepWalk (DW), Node2Vec (N2V), Struc2Vec (S2V) and GraphWave (GW). Best results per GNN architecture and dataset are in \textbf{bold}, second best are \underline{underlined}. Our simple structural attributes often yield competitive or superior performance compared to complex embeddings.}
    \label{tab:metrics_graphs_embeddings}
\end{table*}

\begin{table}[]
    \centering
    \setlength{\tabcolsep}{5pt}
    \begin{tabular}{lccccccc}
    \toprule
     & Feat & Role & Global & DW & N2V & S2V & GW \\
    \midrule
    Texas & 0.48 & 0.03 & 0.02 & 1.10 & 0.21 & 1.69 & 0.04 \\
    Wisconsin & 0.45 & 0.03 & 0.03 & 0.52 & 0.41 & 2.44 & 0.05 \\
    Cornell & 0.37 & 0.02 & 0.02 & 0.22 & 0.20 & 1.57 & 0.03 \\
    Actor & 29.56 & 8.03 & 15.77 & 96.95 & 43.19 & 555.29 & 65.06 \\
    Chameleon & 7.18 & 0.88 & 1.81 & 26.24 & 16.22 & 156.55 & 3.36 \\
    Squirrel & 30.38 & 4.55 & 14.88 & 83.12 & 116.59 & 622.74 & 24.86 \\
    Cora & 6.03 & 1.05 & 1.53 & 41.89 & 15.41 & 124.38 & 12.37 \\
    CiteSeer & 22.22 & 1.66 & 1.87 & 47.10 & 17.37 & 118.41 & 9.82 \\
    USA & 1.26 & 0.30 & 0.51 & 3.34 & 2.35 & 31.30 & 0.93 \\
    Brazil & 0.01 & 0.02 & 0.02 & 0.45 & 0.41 & 1.86 & 0.07 \\
    Europe & 0.10 & 0.08 & 0.10 & 1.09 & 0.94 & 9.88 & 0.15 \\
        \bottomrule
    \end{tabular}
    \caption{Computational time (in seconds) for node embedding generation (where applicable) and subsequent $k$-NN and $\epsilon$-ball graph construction using various attribute types (original node features, proposed role-based, proposed global-based, and learned embeddings) on selected datasets. The results highlight the efficiency of our proposed low-dimensional, graph theory-based structural attributes for graph construction compared to computationally intensive embedding generation or graph construction from high-dimensional raw features, an effect particularly noticeable on larger graphs.}
    \label{tab:comp_times_embs}
\end{table}

\begin{table*}[t]
    \centering
    \setlength{\tabcolsep}{2pt}
    \begin{tabular}{l|c|c|c|c|c|c|c|c|c|c|c}
    \toprule
     & Texas & Wisconsin & Cornell & Actor & Chameleon & Squirrel & Cora & CiteSeer & USA & Europe & Brazil \\
    \midrule
    GCN & 49.74{\scriptsize\! $\pm\!\!$ 1.65} & 42.88{\scriptsize\! $\pm\!\!$ 1.03} & 42.63{\scriptsize\! $\pm\!\!$ 1.28} & 28.49{\scriptsize\! $\pm\!\!$ 0.14} & 39.28{\scriptsize\! $\pm\!\!$ 0.3} & 27.65{\scriptsize\! $\pm\!\!$ 0.21} & 87.76{\scriptsize\! $\pm\!\!$ 0.18} & 77.2{\scriptsize\! $\pm\!\!$ 0.24} & 55.8{\scriptsize\! $\pm\!\!$ 0.25} & 35.37{\scriptsize\! $\pm\!\!$ 0.78} & 40.36{\scriptsize\! $\pm\!\!$ 1.34} \\
    GAT & 52.11{\scriptsize\! $\pm\!\!$ 1.38} & 45.19{\scriptsize\! $\pm\!\!$ 0.94} & 50.26{\scriptsize\! $\pm\!\!$ 1.45} & 29.47{\scriptsize\! $\pm\!\!$ 0.19} & 47.64{\scriptsize\! $\pm\!\!$ 0.35} & 29.94{\scriptsize\! $\pm\!\!$ 0.22} & 86.73{\scriptsize\! $\pm\!\!$ 0.16} & 76.08{\scriptsize\! $\pm\!\!$ 0.22} & 53.15{\scriptsize\! $\pm\!\!$ 0.4} & 41.59{\scriptsize\! $\pm\!\!$ 0.75} & 31.43{\scriptsize\! $\pm\!\!$ 1.38} \\
    \midrule
    gfNN & 52.63{\scriptsize\! $\pm\!\!$ 1.48} & 45.58{\scriptsize\! $\pm\!\!$ 1.01} & 43.42{\scriptsize\! $\pm\!\!$ 1.22} & 28.48{\scriptsize\! $\pm\!\!$ 0.19} & 42.79{\scriptsize\! $\pm\!\!$ 0.23} & 28.87{\scriptsize\! $\pm\!\!$ 0.19} & 87.43{\scriptsize\! $\pm\!\!$ 0.18} & 76.66{\scriptsize\! $\pm\!\!$ 0.18} & 58.61{\scriptsize\! $\pm\!\!$ 0.42} & 45.0{\scriptsize\! $\pm\!\!$ 0.61} & 53.21{\scriptsize\! $\pm\!\!$ 1.45} \\
    FAGCN & 83.16{\scriptsize\! $\pm\!\!$ 0.81} & 82.69{\scriptsize\! $\pm\!\!$ 1.05} & 71.32{\scriptsize\! $\pm\!\!$ 1.43} & 35.8{\scriptsize\! $\pm\!\!$ 0.14} & 46.86{\scriptsize\! $\pm\!\!$ 0.3} & 31.43{\scriptsize\! $\pm\!\!$ 0.27} & 75.15{\scriptsize\! $\pm\!\!$ 0.27} & 73.56{\scriptsize\! $\pm\!\!$ 0.3} & 21.93{\scriptsize\! $\pm\!\!$ 0.31} & 23.05{\scriptsize\! $\pm\!\!$ 0.48} & 22.86{\scriptsize\! $\pm\!\!$ 0.86} \\
    DirGNN & 81.32{\scriptsize\! $\pm\!\!$ 1.21} & 82.69{\scriptsize\! $\pm\!\!$ 0.73} & 70.79{\scriptsize\! $\pm\!\!$ 1.65} & 35.56{\scriptsize\! $\pm\!\!$ 0.17} & 65.31{\scriptsize\! $\pm\!\!$ 0.3} & 48.21{\scriptsize\! $\pm\!\!$ 0.18} & 87.7{\scriptsize\! $\pm\!\!$ 0.17} & 76.66{\scriptsize\! $\pm\!\!$ 0.26} & 60.46{\scriptsize\! $\pm\!\!$ 0.41} & 48.41{\scriptsize\! $\pm\!\!$ 0.64} & 55.71{\scriptsize\! $\pm\!\!$ 1.4} \\
    MixHop & 80.79{\scriptsize\! $\pm\!\!$ 0.9} & 82.69{\scriptsize\! $\pm\!\!$ 0.84} & \underline{78.68{\scriptsize\! $\pm\!\!$ 0.93}} & 34.84{\scriptsize\! $\pm\!\!$ 0.14} & 53.19{\scriptsize\! $\pm\!\!$ 0.29} & 35.61{\scriptsize\! $\pm\!\!$ 0.3} & \textbf{88.36{\scriptsize\! $\pm\!\!$ 0.18}} & 76.77{\scriptsize\! $\pm\!\!$ 0.21} & 53.49{\scriptsize\! $\pm\!\!$ 0.31} & 46.83{\scriptsize\! $\pm\!\!$ 0.64} & 53.57{\scriptsize\! $\pm\!\!$ 1.29} \\
    SSGC & 50.26{\scriptsize\! $\pm\!\!$ 1.61} & 47.5{\scriptsize\! $\pm\!\!$ 0.75} & 42.37{\scriptsize\! $\pm\!\!$ 1.1} & 29.39{\scriptsize\! $\pm\!\!$ 0.17} & 39.65{\scriptsize\! $\pm\!\!$ 0.3} & 27.69{\scriptsize\! $\pm\!\!$ 0.2} & 87.89{\scriptsize\! $\pm\!\!$ 0.16} & 77.31{\scriptsize\! $\pm\!\!$ 0.24} & 55.55{\scriptsize\! $\pm\!\!$ 0.24} & 36.83{\scriptsize\! $\pm\!\!$ 0.91} & 40.0{\scriptsize\! $\pm\!\!$ 1.47} \\
    H2GCN & 81.58{\scriptsize\! $\pm\!\!$ 1.15} & 83.46{\scriptsize\! $\pm\!\!$ 0.71} & 74.21{\scriptsize\! $\pm\!\!$ 1.42} & 36.25{\scriptsize\! $\pm\!\!$ 0.16} & 63.06{\scriptsize\! $\pm\!\!$ 0.28} & 47.01{\scriptsize\! $\pm\!\!$ 0.22} & \underline{88.29{\scriptsize\! $\pm\!\!$ 0.18}} & 76.72{\scriptsize\! $\pm\!\!$ 0.21} & 52.14{\scriptsize\! $\pm\!\!$ 0.42} & 50.12{\scriptsize\! $\pm\!\!$ 0.66} & 31.79{\scriptsize\! $\pm\!\!$ 1.33} \\
    \midrule
    SG-GCN$_N$ & 82.11{\scriptsize\! $\pm\!\!$ 1.15} & 82.69{\scriptsize\! $\pm\!\!$ 0.94} & 71.32{\scriptsize\! $\pm\!\!$ 1.39} & \underline{36.76{\scriptsize\! $\pm\!\!$ 0.16}} & 66.2{\scriptsize\! $\pm\!\!$ 0.31} & 53.53{\scriptsize\! $\pm\!\!$ 0.22} & 86.14{\scriptsize\! $\pm\!\!$ 0.21} & \textbf{77.69{\scriptsize\! $\pm\!\!$ 0.28}} & \textbf{65.13{\scriptsize\! $\pm\!\!$ 0.34}} & 52.07{\scriptsize\! $\pm\!\!$ 0.48} & 56.79{\scriptsize\! $\pm\!\!$ 1.7} \\
    SG-GCN & 83.16{\scriptsize\! $\pm\!\!$ 1.34} & 83.85{\scriptsize\! $\pm\!\!$ 0.83} & 72.11{\scriptsize\! $\pm\!\!$ 1.46} & 36.45{\scriptsize\! $\pm\!\!$ 0.18} & 67.01{\scriptsize\! $\pm\!\!$ 0.34} & 59.59{\scriptsize\! $\pm\!\!$ 0.27} & 86.89{\scriptsize\! $\pm\!\!$ 0.15} & \underline{77.57{\scriptsize\! $\pm\!\!$ 0.24}} & 62.86{\scriptsize\! $\pm\!\!$ 0.32} & 46.83{\scriptsize\! $\pm\!\!$ 0.53} & 53.57{\scriptsize\! $\pm\!\!$ 1.49} \\
    SG-GCN$_L$ & \underline{83.95{\scriptsize\! $\pm\!\!$ 0.95}} & 78.27{\scriptsize\! $\pm\!\!$ 0.85} & 72.11{\scriptsize\! $\pm\!\!$ 1.25} & 36.14{\scriptsize\! $\pm\!\!$ 0.19} & \textbf{74.17{\scriptsize\! $\pm\!\!$ 0.34}} & \textbf{69.55{\scriptsize\! $\pm\!\!$ 0.18}} & 87.81{\scriptsize\! $\pm\!\!$ 0.16} & 76.15{\scriptsize\! $\pm\!\!$ 0.22} & \underline{64.20{\scriptsize\! $\pm\!\!$ 0.31}} & \underline{53.17{\scriptsize\! $\pm\!\!$ 0.55}} & \underline{63.21{\scriptsize\! $\pm\!\!$ 1.38}} \\
    SG-FBGNN$_N$ & 81.84{\scriptsize\! $\pm\!\!$ 1.12} & \underline{85.19{\scriptsize\! $\pm\!\!$ 0.8}} & \textbf{79.47{\scriptsize\! $\pm\!\!$ 0.74}} & 32.59{\scriptsize\! $\pm\!\!$ 0.1} & 62.58{\scriptsize\! $\pm\!\!$ 0.47} & 44.11{\scriptsize\! $\pm\!\!$ 0.21} & 81.73{\scriptsize\! $\pm\!\!$ 0.25} & 75.76{\scriptsize\! $\pm\!\!$ 0.34} & 40.5{\scriptsize\! $\pm\!\!$ 0.66} & 47.32{\scriptsize\! $\pm\!\!$ 0.77} & 53.57{\scriptsize\! $\pm\!\!$ 1.77} \\
    SG-FBGNN & 83.16{\scriptsize\! $\pm\!\!$ 1.1} & 85.0{\scriptsize\! $\pm\!\!$ 0.93} & 76.58{\scriptsize\! $\pm\!\!$ 1.03} & 36.31{\scriptsize\! $\pm\!\!$ 0.13} & 67.49{\scriptsize\! $\pm\!\!$ 0.28} & 53.25{\scriptsize\! $\pm\!\!$ 0.34} & 86.14{\scriptsize\! $\pm\!\!$ 0.25} & 76.3{\scriptsize\! $\pm\!\!$ 0.3} & 50.88{\scriptsize\! $\pm\!\!$ 0.41} & 46.95{\scriptsize\! $\pm\!\!$ 0.71} & 46.79{\scriptsize\! $\pm\!\!$ 1.81} \\
    SG-FBGNN$_L$ & \textbf{84.21{\scriptsize\! $\pm\!\!$ 0.88}} & \textbf{85.19{\scriptsize\! $\pm\!\!$ 0.76}} & 76.84{\scriptsize\! $\pm\!\!$ 0.89} & \textbf{37.07{\scriptsize\! $\pm\!\!$ 0.13}} & \underline{71.35{\scriptsize\! $\pm\!\!$ 0.3}} & \underline{63.63{\scriptsize\! $\pm\!\!$ 0.19}} & 86.78{\scriptsize\! $\pm\!\!$ 0.17} & 76.65{\scriptsize\! $\pm\!\!$ 0.21} & 63.28{\scriptsize\! $\pm\!\!$ 0.35} & \textbf{53.17{\scriptsize\! $\pm\!\!$ 0.55}} & \textbf{63.93{\scriptsize\! $\pm\!\!$ 1.6}} \\
    \bottomrule
    \end{tabular}
    \caption{Node classification accuracy (\%) comparing our SG-GNN architectures with state-of-the-art baselines across all datasets. SG-GNN$_N$ refers to the node-specific adaptive single-layer model, SG-GNN is the global adaptive single-layer model, and SG-GNN$_L$ is the multi-layer ($L=2$) adaptive model. Each is implemented with both GCN and FBGNN base layers. Best results per dataset are in \textbf{bold}, second best are \underline{underlined}.}
        \label{tab:metrics_sggnn}
\end{table*}

This section presents our empirical evaluation. 
Section~\ref{Ss:datasets} introduces the datasets, after which Section~\ref{Ss:eval_proposed_graphs} quantifies the gains obtained when graphs produced by our structure-based neighbor discovery (Section~\ref{S:neighbor_discovery}) are integrated into standard GNN backbones. 
Section~\ref{Ss:eval_sggnn_architectures} then benchmarks the complete SG-GNN architectures (Section~\ref{S:sg-gnn}) against state-of-the-art baselines and analyzes the learned coefficients of the interpretable architecture. 

\subsection{Datasets} \label{Ss:datasets}

To comprehensively evaluate the performance of our proposed SG-GNN framework, we conduct experiments on eleven benchmark graph datasets.
These include the webpage co-occurrence networks Texas, Wisconsin, and Cornell~\cite{pei2020geom, sen2008collective}; the actor co-occurrence network Actor~\cite{pei2020geom}; the Wikipedia page-link networks Chameleon and Squirrel~\cite{pei2020geom}; the citation networks Cora~\cite{mccallum2000cora} and CiteSeer~\cite{giles98citeseer}; and air traffic networks in USA, Brazil, and Europe~\cite{ribeiro2017struc2vec}.
As detailed in Table~\ref{tab:smoothness}, the original graph $\ccalG$ for these datasets exhibit a wide spectrum of homophily levels: Texas, Wisconsin, Cornell, Actor, Chameleon, and Squirrel are highly heterophilic with edge homophily ratios below 0.3; USA, Europe, and Brazil present intermediate levels (between 0.3 and 0.7); while Cora and CiteSeer are strongly homophilic (above 0.7).
These datasets span a range of sizes, node feature dimensionalities, and class numbers.
We follow standard public data splits for training, validation, and testing where available, or use commonly adopted random splits otherwise.

\subsection{Evaluation of the proposed graphs} \label{Ss:eval_proposed_graphs}

We first evaluate the performance of GNNs using the alternative graph topologies introduced in Section~\ref{S:neighbor_discovery}. We construct $k$-NN $\mathcal{G}_k$ and $\epsilon$-ball $\mathcal{G}_\epsilon$ graphs using various attribute types: our proposed interpretable role-based and global-based structural attributes, original node features $\bbX$, and several established learned node embedding techniques, DeepWalk~\cite{perozzi2014deepwalk}, Node2Vec~\cite{grover2016node2vec}, Struc2Vec~\cite{ribeiro2017struc2vec}, and GraphWave~\cite{donnat2018learning}. 

Table~\ref{tab:metrics_graphs_embeddings} summarizes the node classification accuracies achieved on selected datasets using these constructed graphs.
We apply two types of GNN layers: a standard GCN layer~\cite{kipf17gnns} and an FBGNN layer employing a generic graph filter [cf.~\eqref{eq:gcn_gf_layer}], which allows us to evaluate alternative graphs for both homophily-dependent GNNs and those adaptable to heterophilic data.
In comparison to graphs built from more complex learned embeddings, we observe competitive performance for graphs $\mathcal{G}^{\rm nn}_{k, \text{role}}, \mathcal{G}^{\rm ball}_{\epsilon, \text{role}}$ derived from role-based attributes; $\mathcal{G}^{\rm nn}_{k, \text{glob}}, \mathcal{G}^{\rm ball}_{\epsilon, \text{glob}}$ from global attributes; and $\mathcal{G}^{\rm nn}_{k, \text{feat}}, \mathcal{G}^{\rm ball}_{\epsilon, \text{feat}}$ from the original node features.
For example, on Wisconsin with GCN, $\ccalG^{\rm ball}_{\epsilon, \text{feat}}$ leads, with $\ccalG^{\rm ball}_{\epsilon, \text{role}}$ and $\ccalG^{\rm ball}_{\epsilon, \text{S2V}}$ also performing strongly. Similarly, for Actor with FBGNN, $\ccalG^{\rm ball}_{\epsilon, \text{role}}$ achieves the best result. 
Thus, interpretable structural information may improve GNN performance and preclude the need to use computationally intensive embedding methods.

Beyond classification performance, a significant advantage of our proposed structural attributes lies in their computational efficiency. 
While detailed timings are presented in Table~\ref{tab:comp_times_embs}, we highlight here that our graph theory-based role and global features are inherently fast to compute. 
In contrast, methods like DeepWalk, Node2Vec, Struc2Vec, and GraphWave involve substantial computational overhead for generating the embeddings. 
Indeed, constructing graphs with Struc2Vec yields models that are conceptually similar to that of~\cite{suresh2021breaking}, albeit with a far simpler graph construction method, yet the complexity of performing Struc2Vec causes $\ccalG_{k,\text{S2V}}^{\rm nn}$ to consistently incur the greatest computational time in Table~\ref{tab:comp_times_embs}.
Even when using raw node features for $\mathcal{G}^{\rm nn}_{k, \text{feat}}$, which require no explicit embedding computation time, the subsequent step of building $k$-NN or $\epsilon$-graphs can be time-consuming if the original feature dimensionality $M$ is high, as distance computations become expensive. 
Our proposed role-based and global features are typically low-dimensional, leading to faster graph construction. 
This efficiency becomes particularly crucial for larger graphs, such as Actor or Squirrel, where the cost of obtaining complex embeddings or high-dimensional feature-based graph construction can become prohibitive. 
Thus, our structure-guided approach offers a compelling balance of strong performance and computational tractability.

\subsection{Evaluation of SG-GNN} \label{Ss:eval_sggnn_architectures}

We now evaluate the performance of our proposed SG-GNN architectures against two established GNN baselines and six specialized methods for heterophily.
As foundational benchmarks, we include the GCN~\cite{kipf17gnns} and the GAT~\cite{velickovic2018graph}, which are widely adopted but primarily designed for homophilic graphs.
More central to our evaluation, we compare against several methods specifically designed or proved to be effective for heterophilic graphs.
These include gfNN~\cite{nt2020revisiting}, FAGCN~\cite{bo2021beyond}, DirGNN~\cite{rossi2024edge}, MixHop~\cite{abu2019mixhop}, SSGC~\cite{zhu2021simple} and H2GCN~\cite{zhu2020beyond}.

For the proposed SG-GNN, we consider three main variants of our SG-GNN:
(i) SG-GNN, the single-layer SG-GNN with global adaptive graph weights as introduced in Section~\ref{S:sg-gnn};
(ii) SG-GNN$_N$, the single-layer SG-GNN incorporating node-specific aggregation weights as described in Section~\ref{Ss:node-specific};
and (iii) SG-GNN$_L$, the multi-layer SG-GNN architecture stacking $L$ layers of the adaptive graph aggregation, as detailed in Section~\ref{Ss:full_sg-gnn} (for our experiments, we use $L=2$).
Each of these architectures is implemented using both standard GCN layers and the FBGNN layer described in the previous section as the underlying GNN operation per graph. For these SG-GNN models, the set of input graphs comprises the original graph, $k$-NN graphs ($k=3$) constructed from role-based and global structural attributes, and from the embeddings introduced in Section~\ref{Ss:eval_proposed_graphs}.

Table~\ref{tab:metrics_sggnn} presents a comprehensive comparison of node classification accuracy for our SG-GNN framework against state-of-the-art baselines across all eleven datasets. The results highlight the broad effectiveness of our approach. Notably, different variants of our SG-GNN (SG-GNN$_N$, SG-GNN, and SG-GNN$_L$, using either GCN or FBGNN base layers) achieve the best performance on 10 out of the 11 datasets evaluated. This consistent top-tier performance, particularly evident on the more challenging heterophilic datasets, underscores the robustness and utility of adaptively leveraging multiple structurally-informed graph views for node classification.

Across the board, whether using GCN or FBGNN as the base layer, and whether employing the single-layer adaptive versions (SG-GNN$_N$, SG-GNN) or the multi-layer variant (SG-GNN$_L$), our proposed architectures show a strong ability to leverage multiple graph structures effectively. The adaptive weighting schemes allow the models to dynamically emphasize the most informative graph(s) for the task at hand, while the multi-layer approach enables the learning of more complex interactions between features and diverse structural information. These results underscore the benefits of explicitly guiding GNNs with structurally-informed alternative graph topologies, leading to robust performance improvements, especially in scenarios with non-homophilic data.

\begin{figure}[t]
    \centering
    \begin{tikzpicture}[baseline,scale=.55]

\pgfplotsset{colormap={CM}{
    rgb255=(54, 75, 154)
    rgb255=(74, 123, 183)
    rgb255=(110, 166, 205)
    rgb255=(152, 202, 225)
    rgb255=(253, 179, 102)
    rgb255=(246, 126, 75)
    rgb255=(221, 61, 45)
    rgb255=(165, 0, 38)
}}

\begin{groupplot}[
    name=learned_coefs,
    table/col sep=space,
    width=14cm,
    height=7cm,
    group style={group size=1 by 2,
        horizontal sep=1cm,
        vertical sep=4.5cm,},
    enlargelimits=false,
    colorbar horizontal,
    xtick={0,1,2,3,4,5,6,7,8,9,10,11,12,13},
    xticklabels={$\mathcal{G}\quad \;$,$\ccalG^{\rm ball}_{\epsilon, {\rm feat}}$,$\ccalG^{nn}_{k, {\rm feat}}$,$\ccalG^{\rm ball}_{\epsilon, {\rm role}}$,$\ccalG^{nn}_{k, {\rm role}}$,$\ccalG^{\rm ball}_{\epsilon, {\rm glob}}$,$\ccalG^{nn}_{k, {\rm glob}}$,$\ccalG^{\rm ball}_{\epsilon, {\rm DW}}$,$\ccalG^{nn}_{k, {\rm DW}}$,$\ccalG^{\rm ball}_{\epsilon, {\rm N2V}}$,$\ccalG^{nn}_{k, {\rm N2V}}$,$\ccalG^{\rm ball}_{\epsilon, {\rm S2V}}$,$\ccalG^{nn}_{k, {\rm S2V}}$,$\ccalG^{nn}_{k, {\rm GW}}$},
    x tick label style={rotate=225,anchor=west, yshift=-0.1cm, xshift=0.3cm},
    colorbar style={
        xtick={0.0, 0.05, 0.1, 0.15, 0.2, 0.25, 0.3, 0.35, 0.4, 0.45},
        xticklabel style={ 
            /pgf/number format/.cd, 
            fixed,                
            fixed zerofill,       
            precision=2,          
            /tikz/.cd,             
            xshift=0cm,
            yshift=-0.1cm
        }
    },
    ytick={0,1,2,3,4,5,6,7,8,9,10},
    yticklabels={Texas,Wisconsin,Cornell,Actor,Chameleon,Squirrel,Cora,CiteSeer,USA,Europe,Brazil},
    axis on top,
    label style={font=\LARGE},
    tick style={draw=none},
    tick label style={font=\large},
    title style={font=\huge},
    x tick label style={rotate=90}
    ]

    \pgfplotstableread{images/data/coefs-GCN.csv}\matrixA
    \pgfplotstableread{images/data/coefs-FBGNN.csv}\matrixB

\nextgroupplot[title={(a) GCN},xmin=-0.5,xmax=13.5,ymin=-0.5,ymax=10.5]
    \addplot [
        matrix plot,
        colorbar,
        nodes near coords,
        every node near coord/.append style={
            font=\footnotesize,
            /pgf/number format/fixed,        
            /pgf/number format/precision=2,  
            text=white,
            anchor=center
        },
        point meta=explicit,
        mesh/cols=14,  
    ] table [meta=value] {\matrixA};

\nextgroupplot[title={(b) FBGNN},xmin=-0.5,xmax=13.5,ymin=-0.5,ymax=10.5]
    \addplot [
        matrix plot,
        colorbar,
        nodes near coords,
        every node near coord/.append style={
            font=\footnotesize,
            /pgf/number format/fixed,        
            /pgf/number format/precision=2,  
            text=white,
            anchor=center,
        },
        point meta=explicit,
        mesh/cols=14,  
    ] table [meta=value] {\matrixB};

\end{groupplot}
\end{tikzpicture}
    \caption{Learned adaptive coefficients $\alpha_r$ for the single-layer SG-GNN architecture using (a) GCN base layers and (b) FBGNN base layers. Each row represents a dataset, and columns correspond to different input graph types: original graph ($\mathcal{G}$), and $\epsilon$-ball or $k$-NN graphs derived from original features (feat), role-based attributes (role), global-based attributes (global), DeepWalk (DW), Node2Vec (N2V), Struc2Vec (S2V), and GraphWave (GW). Warmer colors indicate higher learned importance for the corresponding graph structure.}
    \label{fig:learned_coefs}
\end{figure}
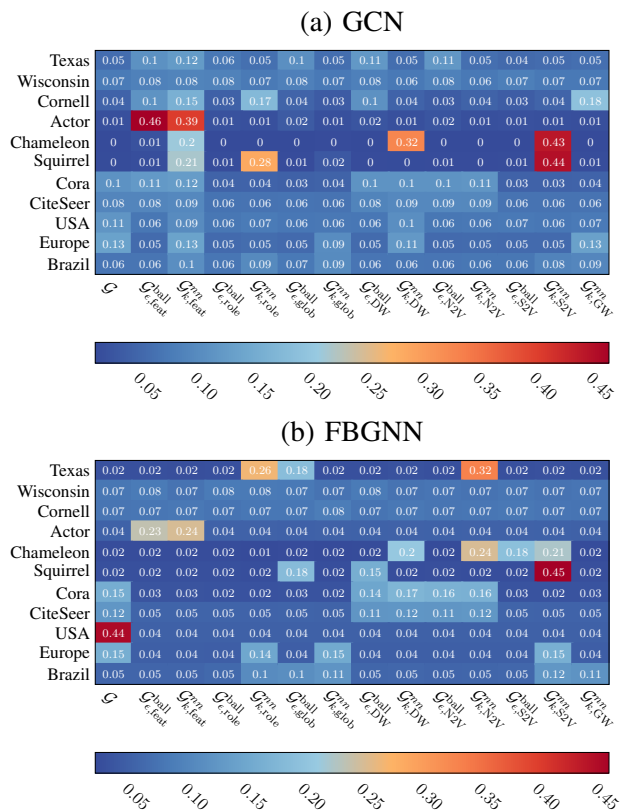

To understand how our SG-GNN adapts to different graph structures, we examine the learned coefficients $\alpha_r$ (Section~\ref{S:sg-gnn}) which, together with the weights of the MLP $\psi$, relate to the importance of each input graph in the architecture. 
Fig.~\ref{fig:learned_coefs} visualizes the coefficients $\alpha_r$ for the single-layer SG-GNN, using a broad set of candidate graphs (original $\mathcal{G}$, and variants from features, role-based/global attributes, and common embeddings) with GCN and FBGNN base layers. 
For datasets with higher $h_{\rm edge}$ in Table~\ref{tab:smoothness}, the model indeed tends to consider the original graph $\ccalG$ among the most informative, along with other graphs that performed individually well in Table~\ref{tab:metrics_graphs_embeddings}.
Moreover, the GCN also incorporates $\ccalG_{k,\text{feat}}^{\rm nn}$ and $\ccalG_{\epsilon,\text{feat}}^{\rm ball}$ for the famously homophilic Cora and Citeseer, where both node features and labels tend to be smooth on $\ccalG$.
Conversely, the more expressive FBGNN allows the SG-GNN to select graphs without being limited to those on which the data is smooth.
For example, the FBGNN achieves the best performance on the highly heterophilic Texas dataset, where weights are significantly higher for $\ccalG_{k,\text{role}}^{\rm nn}$, $\ccalG_{\epsilon,\text{glob}}^{\rm ball}$, and $\ccalG_{k,\text{N2V}}^{\rm nn}$.
This aligns with the increased homophily for Texas on graphs constructed from role-based and global structural attributes in Table~\ref{tab:smoothness} and Fig.~\ref{fig:homophilies}.
This interpretable weighting approach indicates not only the value of combining multiple different graph structures but also, as no single graph type performs best across all datasets, demonstrates the need for adaptive architectures such as our SG-GNN.



\section{Conclusion}
\label{S:conc}

In this work, we addressed the challenge of applying GNNs to heterophilic data by proposing a structure-guided neighbor discovery approach. We showed, both theoretically and empirically, that constructing new graph topologies based on node structural attributes can significantly enhance data homophily and subsequent GNN performance. Our key contribution is SG-GNN, an adaptive architecture that learns to effectively combine the original graph with these novel, structurally-informed graphs. Experiments show that SG-GNN, leveraging simple and interpretable graph-theoretic features, achieves highly competitive performance on various benchmark datasets, particularly those exhibiting heterophily, while also offering insights into which structural aspects are most beneficial for the learning task.

Future research directions include exploring a broader range of structural attributes and developing methods to automatically learn or select optimal structural attributes for graph construction. Extending the SG-GNN framework to other graph learning tasks, such as link prediction and graph classification, also presents a promising avenue. Furthermore, a deeper investigation into the interplay between node-level and graph-level adaptive weighting could yield even more fine-grained interpretability and performance gains. These directions promise to further advance the capabilities of GNNs in handling complex and diverse graph-structured data.


\bibliographystyle{IEEEtran}
\bibliography{myIEEEabrv,citations}

\appendices

\section{Structural Feature Definitions}\label{app:struc_features}

Below we provide a brief definition for each of the structural attributes computed for nodes based on the original graph $\mathcal{G}=(\ccalV,\ccalE)$ with adjacency matrix $\bbA$. The egonet of a node $i$ is defined as the subgraph induced by node $i$ and its immediate neighbors $\ccalN(i)$.

\subsection{Global Structural Attributes}
These features capture a node's properties in relation to the entire graph structure~\cite{newman2018networks}.
\begin{enumerate}
    \item \textbf{Eccentricity:} Maximum shortest path distance from the node to any other node within its connected component.
    \item \textbf{PageRank Centrality:} Node importance score based on the link structure, as per the PageRank algorithm.
    \item \textbf{Eigenvector Centrality:} Node influence score, where a high score means the node is connected to other highly central nodes.
    \item \textbf{Betweenness Centrality:} Extent to which a node lies on shortest paths between other pairs of nodes.
    \item \textbf{Closeness Centrality:} Inverse of the average shortest path distance from the node to all other reachable nodes in the graph.
    \item \textbf{Katz Centrality:} Node influence measured by the sum of attenuated paths ending at the node; nodes are influential if they are connected to many nodes or to other influential nodes.
    \item \textbf{Core Number:} The $k$-value of the highest $k$-core (maximal subgraph where every node has degree at least $k$) that the node belongs to.
\end{enumerate}

\subsection{Role-Based Structural Attributes}
These features characterize a node's local connectivity patterns and its role within its immediate neighborhood (egonet). These set of features are inspired from those used in~\cite{guo20role}.
\begin{enumerate}
    \item \textbf{Degree:} Number of direct neighbors (edges) of the node.
    \item \textbf{Egonet Edge Sum:} Sum of all entries in the adjacency matrix of the node's egonet (i.e., twice the number of edges within the egonet).
    \item \textbf{Egonet Total Degree:} Sum of the degrees (in the original graph) of all nodes that belong to the node's egonet.
    \item \textbf{Egonet Internal Connectivity Proportion:} Ratio of twice the number of internal egonet edges to the sum of degrees of all egonet nodes; measures how much of the egonet's connectivity is self-contained.
    \item \textbf{Egonet External Connectivity Proportion:} $1 - (\text{Egonet Internal Connectivity Proportion})$; measures the proportion of egonet connectivity that links to nodes outside the egonet.
    \item \textbf{Triangle Participation (scaled):} Twice the number of distinct triangles the node is a part of, computed as the diagonal entries of $\bbA^3$.
    \item \textbf{Scaled Local Clustering Coefficient:} Twice the standard local clustering coefficient, measuring the density of connections among a node's immediate neighbors.
\end{enumerate}

\section{Proof of Theorem~\ref{thm:error_bound}}
\label{app:thm_proof}

First, observe that since $\bbX^*$ recovers $\bby$ according to Definition~\ref{def:recover}, then $\bbX^* = \hArws \bbX^*$.
Moreover, because $\bbTheta^{(1)}$ and $\bbTheta^{(2)}$ perform column-wise linear transformations and $\sigma_1$ and $\sigma_2$ are element-wise operations, the matrices $\sigma_1( \bbX^* \bbTheta^{(1)} )$ and $\sigma_2(\sigma_1( \bbX^*\bbTheta^{(1)} ) \bbTheta^{(2)} )$ also recover $\bby$.
Thus, we have that
\alna{
    \bbZ^*
    &~:=~&
    \sigma_2( \hArws 
    \sigma_1( \hArws \bbX^* \bbTheta^{(1)} )
    \bbTheta^{(2)})
&\nonumber\\&
    &~=~&
    \sigma_2( \hArws 
    \sigma_1( \bbX^* \bbTheta^{(1)} )
    \bbTheta^{(2)})
&\nonumber\\&
    &~=~&
    \sigma_2( 
    \sigma_1( \bbX^* \bbTheta^{(1)} )
    \bbTheta^{(2)}),
\nonumber}
so $\bbZ^*$ also recovers $\bby$.

Next, we bound the difference between $\bbZ^*$ and $\hbZ$. Consider
\alna{
    \| \bbZ^* - \hbZ \|_F
    &~\leq~&
    \| \bbZ^* - \tbZ \|_F
    +
    \| \tbZ - \hbZ \|_F,
\label{eq:Z_err_tri_ineq}}
where $\tbZ := \Phi( \bbX; \bbA^*, \bbTheta )$.
We bound the first term of~\eqref{eq:Z_err_tri_ineq} as
\alna{
    \| \bbZ^* - \tbZ \|_F
    &\leq&
    \rho_2 
    \left\| 
        \hArws \Big[
            \sigma_1(\hArws \bbX^* \bbTheta^{(1)})
            -
            \sigma_1(\hArws \bbX \bbTheta^{(1)})
        \Big]\!
    \right\|_F
&\nonumber\\&
    &\leq&
    \rho_2 
    \left\| 
        \sigma_1(\hArws \bbX^* \bbTheta^{(1)})
        -
        \sigma_1(\hArws \bbX \bbTheta^{(1)})
    \right\|_F,
\nonumber}
where the first inequality arises by the definition of $\rho_2$ and the nonexpansiveness of $\sigma_2$, i.e., $\| \sigma_2(\bbX_1)-\sigma_2(\bbX_2) \|_F \leq \|\bbX_1-\bbX_2\|_F$, and the second by the fact that $\| \hArws \|_1 = 1$.
By the analogous definitions of $\rho_1$ and $\sigma_1$, we further have
\alna{
    \| \bbZ^* - \tbZ \|_F
    &~\leq~&
    \rho_1 \rho_2 
    \left\| 
        \hArws
        (\bbX^* - \bbX)
    \right\|_F.
\label{eq:term1_bnd}}
Then, given the degree vector $\hbd^* := (\bbA^*+\bbI)\bbone$ and our assumption bounding the noise in $\bbX$,
\begin{align}
    \| [\hArws(\bbX^* - \bbX)]_{i,:} \|_2
    =
    \Big\|
        \frac{1}{ \hat{d}_i^* }\!
        \sum_{j=1}^N
        [\bbI + \bbA^*]_{ij}
        [\bbX^* - \bbX]_{i,:}
    \Big\|_2
\nonumber\\
    \hspace{0.8cm}~\leq~
    \frac{1}{ \hat{d}_i^* }
    \sum_{j=1}^N
    [\bbI + \bbA^*]_{ij}
    \| [\bbX^* - \bbX]_{i,:} \|_2
    ~\leq~
    \alpha,
\nonumber
\end{align}
so we have that
\alna{
    \| \bbZ^* - \tbZ \|_F
    &~\leq~&
    \rho_1 \rho_2
    \alpha \sqrt{N}.
\nonumber}

For the second term of~\eqref{eq:Z_err_tri_ineq}, we again exploit the definitions of $\rho_2$ and $\sigma_2$ for
\alna{
    \| \tbZ - \hbZ \|_F
    &\leq&
    \rho_2 
    \left\|
        \hArws \sigma_1(\hArws \bbX \bbTheta^{(1)})
        -
        \hArw \sigma_1(\hArw \bbX \bbTheta^{(1)})
    \right\|_F
&\nonumber\\&
    &=&
    \rho_2
    \left\|
        \hArws \sigma_1(\hArws \bbX \bbTheta^{(1)})
        -
        \hArws \sigma_1(\hArw \bbX \bbTheta^{(1)})
    \right.
&\nonumber\\&
    && \quad
    \left.
        +
        \hArws \sigma_1(\hArw \bbX \bbTheta^{(1)})
        -
        \hArw \sigma_1(\hArw \bbX \bbTheta^{(1)})
    \right\|_F
&\nonumber\\&
    &\leq&
    \rho_2 
    \left\|
        \hArws \big( \sigma_1(\hArws \bbX \bbTheta^{(1)})
        -
        \sigma_1(\hArw \bbX \bbTheta^{(1)}) \big)
    \right\|_F
&\nonumber\\&
    && \quad
    +
    \rho_2 
    \left\|
        \left( \hArws - \hArw \right) 
        \sigma_1(\hArw \bbX \bbTheta^{(1)}) \big)
    \right\|_F,
\label{eq:term2_bnd_pt1}}
where we apply the triangle inequality for~\eqref{eq:term2_bnd_pt1}.
Similarly to the steps for~\eqref{eq:term1_bnd}, we apply the nonexpansiveness of $\sigma_1$, the definition of $\rho_1$, and $\| \hArw \|_1 = 1$ twice to get
\alna{
    \| \tbZ - \hbZ \|_F
    &~\leq~&
    \rho_1\rho_2
    \left\| \left( \hArws - \hArw \right) \bbX \right\|_F
&\nonumber\\&
    &&
    +
    \rho_2
    \left\| \left( \hArws - \hArw \right) 
    \sigma_1(\hArws\bbX\bbTheta^{(1)})
    \right\|_F
&\nonumber\\&
    &~\leq~&
    \rho_1 \rho_2
    \left\| \hArws - \hArw \right\|_F
    \| \bbX \|_F
&\nonumber\\&
    &&
    +
    \rho_2
    \left\| \hArws - \hArw \right\|_F
    \left\| \sigma_1(\hArws\bbX\bbTheta^{(1)}) \right\|_F
&\nonumber\\&
    &~\leq~&
    2\rho_1 \rho_2
    \left\| \hArws - \hArw \right\|_F
    \| \bbX \|_F.
\nonumber}
Next, we consider the diagonal matrices $\hbD = \diag((\bbA+\bbI)\bbone)$ and $\hbD^* = \diag((\bbA^* + \bbI) \bbone)$ such that $\hArw = \hbD^{-1} \hbA$ and $\hArws = (\hbD^*)^{-1} \hbA^*$.
Then, we have that
\alna{
    \| \tbZ - \hbZ \|_F
    &~\leq~&
    2 \rho_1\rho_2
    \| ( \hArws - \hArw ) \|_F
    \| \bbX \|_F
&\nonumber\\&
    &~=~&
    2 \rho_1\rho_2
    \| \bbX \|_F
    \big\| \hArws - \hbD^{-1}\hbA^* 
    \big.
&\nonumber\\&
    &&
    \big.
    \qquad\qquad\qquad
    + \hbD^{-1}\hbA^* - \hArw \big\|_F
&\nonumber\\&
    &~\leq~&
    2 \rho_1 \rho_2
    \| \bbX \|_F
    \Big(
        \| \hArws - \hbD^{-1}\hbA^*  \|_F
        \Big.
&\nonumber\\&
    &&
        \Big.
        \qquad\qquad\qquad
        +
        \| \hbD^{-1} (\hbA^*-\hbA)  \|_F
    \Big).
\nonumber}
Then, we replace $\hbA^* = \hbD^* \hArws$ in the first term of the above upper bound and $\bbDelta = \bbA - \bbA^* = \hbA - \hbA^*$ in the second term for
\alna{
    \| \tbZ - \hbZ \|_F
    &~\leq~&
    2 \rho_1 \rho_2
    \Big( \| (\bbI - \hbD^{-1}\hbD^*) \hArws \|_F
        + \| \hbD^{-1} \bbDelta \|_F
    \Big)
&\nonumber\\&
    &~=~&
    2 \rho_1 \rho_2
    \Big( \| \hbD^{-1} (\hbD - \hbD^*) \hArws \|_F
        + \| \hbD^{-1} \bbDelta \|_F
    \Big)
&\nonumber\\&
    &~\leq~&
    2 \rho_1 \rho_2
    \Big( \| (\hbD - \hbD^*) \hArws \|_F
        + \| \bbDelta \|_F
    \Big)
&\nonumber\\&
    &~\leq~&
    2 \rho_1 \rho_2
    \Big( \| \hbD - \hbD^* \|_F
        + \| \bbDelta \|_F
    \Big),
\nonumber}
where the last inequality holds by $\| \hArws \|_1 = 1$.
By the definitions of $\hbD$, $\hbD^*$, and $\bbDelta$,
\alna{
    \| \tbZ - \hbZ \|_F
    &~\leq~&
    2 \rho_1 \rho_2
    \Big(
        \| \diag( \bbDelta \bbone ) \|_F
        +
        \| \bbDelta \|_F
    \Big)
&\nonumber\\&
    &~=~&
    2 \rho_1 \rho_2
    \Big(
        \| \bbDelta\bbone \|_2
        +
        \| \bbDelta \|_F
    \Big)
&\nonumber\\&
    &~\leq~&
    2 \rho_1 \rho_2
    \Big(
        \sqrt{N}
        \| \bbDelta \|_F
        +
        \| \bbDelta \|_F
    \Big).
\label{eq:term2_bnd}}
Finally, summing the upper bounds in~\eqref{eq:term1_bnd} and~\eqref{eq:term2_bnd} yields the error bound in~\eqref{eq:err_bnd}, as desired.

\section{Proof of Proposition~\ref{prop:sbm}}
\label{app:prop1_proof}

Let $\ccalG_r$ be the $r$-th graph in the given set $\{\ccalG_r\}_{r=1}^R$.
Then, let $\beta_r$ denote the event that there are no edges between nodes with different labels, that is
\alna{
    \beta_r
    &~:=~&
    \{ 
    y_i = y_j 
    ~\forall\, (i,j) \in \ccalE_r
    \}.
\nonumber}
To determine the likelihood of $\beta_r$ occurring, first we consider the number of pairs of nodes that belong to different classes,
\alna{
    \left| \left\{
        (i,j) \in \ccalV \times \ccalV
        :\,
        y_i\neq y_j
    \right\} \right|
    &~=~&
    \frac{N(N-1)}{2}
    -
    \frac{CK(K-1)}{2}
&\nonumber\\&
    &~=~&
    \frac{N^2 (C-1)}{2C},
\nonumber}
where we use the assumption of equally sized classes $K = N/C$.
Then, since edge likelihoods are independent for all pairs of nodes, we have that
\alna{
    \mbP[\beta_r]
    &~\geq~&
    (1-q)^{ \frac{N^2 (C-1) }{ 2C } }.
\label{eq:prob_ideal_r}}
We are interested in the likelihood that at least one of the $R$ graphs does not have false positive edges, that is, $\beta := \bigcup_{r=1}^R \beta_r$.
Since the graphs are assumed to be independent, we have that
\alna{
    1-\mbP[\beta]
    &~=~&
    \mbP\left[
        \bigcap_{r=1}^R \bar{\beta}_r
    \right]
    ~=~
    \left(
        1-\mbP[\beta_r]
    \right)^R
&\nonumber\\&
    &~\leq~&
    \left(
        1 - (1-q)^{\frac{N^2 (C-1)}{2C}}
    \right)^R,
\nonumber}
where $\bar{\beta}_r$ denotes the complement of $\beta_r$.
Taking the complement of both sides of the inequality yields the probability bound in~\eqref{eq:sbm_prob}, as desired.

\section{Proof of Proposition~\ref{prop:sbm_node}}
\label{app:prop2_proof}

Consider a specific node $i\in\ccalV$ and graph $\ccalG_r$ from the given set $\{\ccalG_r\}_{r=1}^R$.
We define the event
\alna{
    \gamma_{i,r}
    &~:=~&
    \left\{
        y_i = y_j
        ~:\, 
        j \in \ccalN_r(i)
    \right\},
\nonumber}
which occurs when all neighbors of the $i$-th node of graph $\ccalG_r$ share the same label $y_i$.
The probability of $\gamma_{i,r}$ is bounded below by
\alna{
    \mbP[\gamma_{i,r}]
    &~\geq~&
    (1-q)^{ \frac{N(C-1)}{C} }.
\nonumber}
We first want to find the probability that there exists at least one of the $R$ graphs for which all neighbors of the $i$-th node share its label $y_i$, that is, $\gamma_i := \bigcup_{r=1}^R \gamma_{i,r}$.
Similarly to the proof of Proposition~\ref{prop:sbm}, let $\bar{\gamma}_{i,r}$ be the complement of $\gamma_{i,r}$.
We then have that
\alna{
    1 - \mbP[\gamma_i]
    &~=~&
    \mbP\left[
        \bigcap_{r=1}^R \bar{\gamma}_{i,r}
    \right]
    ~\leq~
    \left(
        1 - (1-q)^{ \frac{N(C-1)}{C} }
    \right)^R.
\nonumber}
We can then obtain the probability that $\gamma_i$ holds for all $i\in\ccalV$, $\gamma := \bigcap_{i=1}^N \gamma_i$.
Since all graphs are assumed to be independent,
\alna{
    \mbP\left[\gamma\right]
    &~=~&
    \mbP\left[\bigcap_{i=1}^N\gamma_i\right]
    \,=\,
    \prod_{i=1}^N
    \mbP\left[ \gamma_i \right]
&\nonumber\\&
    &~\geq~&
    \left[
    1-
    \left(
        1 - (1-q)^{ \frac{N(C-1)}{C} }
    \right)^R
    \right]^N,
\nonumber}
which is the result in~\eqref{eq:sbm_prob_node}.

\end{document}